\documentclass{article}

\usepackage[main, final]{neurips_2026}

\usepackage[dvipsnames]{xcolor}
\usepackage[utf8]{inputenc} 
\usepackage[T1]{fontenc}    
\usepackage[colorlinks=true, citecolor=RoyalBlue]{hyperref}       
\usepackage{url}            
\usepackage{booktabs}       
\usepackage{amsfonts}       
\usepackage{nicefrac}       
\usepackage{microtype}      
\usepackage{xcolor}         
\usepackage{enumitem}
\usepackage{amsmath}
\usepackage{multirow}
\usepackage{booktabs} 
\usepackage{makecell}  
\usepackage[table]{xcolor}
\usepackage{graphicx}
\usepackage[most]{tcolorbox}

\author{
{\bfseries Junxi Wang\textsuperscript{1,2},
Te Sun\textsuperscript{1},
Jiayi Zhu\textsuperscript{1},
Chen Zhang\textsuperscript{4}}
Siyuan Li\textsuperscript{5},
Xuyang Liu\textsuperscript{6},\\[0.1em]
{\bfseries Zichen Wen\textsuperscript{1,3},
Xiaobing Tu\textsuperscript{7},
Jinkui Ren\textsuperscript{7},
Xiantao Zhang\textsuperscript{7},
Ziqi Yuan\textsuperscript{8},
Linfeng Zhang\textsuperscript{1}\thanks{Corresponding author.}}\\[0.1em]
\textsuperscript{1}Shanghai Jiao Tong University,\hspace{0.05cm}
\textsuperscript{2}Fudan University,\hspace{0.05cm}
\textsuperscript{3}Shanghai AI Lab,\hspace{0.05cm}
\textsuperscript{4}Nanjing University\\[0.1em]
\textsuperscript{5}HIT,\hspace{0.05cm}
\textsuperscript{6}The Hong Kong Polytechnic University,\hspace{0.05cm}
\textsuperscript{7}Alibaba Group,\hspace{0.05cm}
\textsuperscript{8}Tsinghua University\\[0.1em]
\texttt{junxiwang182@gmail.com},\hspace{0.2cm}\texttt{zhanglinfeng@sjtu.edu.cn}
}

\makeatletter
\newif\if@appendixtoc
\@appendixtocfalse
\newcommand{\appendixTOCstart}{\global\@appendixtoctrue}

\let\old@contentsline\contentsline
\renewcommand{\contentsline}[4]{%
  \if@appendixtoc
    \old@contentsline{#1}{#2}{#3}{#4}%
  \fi
}
\makeatother

\title{MemForest: Efficient Agent Memory Management via EventTree Partitioning and Progressive Merging}

\begin{document}

\maketitle

\begin{abstract}
    Agent memory systems have demonstrated significant potential in tasks such as long-term dialogue, personalized assistants, and video understanding. However, as inference progresses, continuously accumulated memory imposes substantial storage and retrieval burdens. To address this issue, we propose \textbf{MemForest}, a general memory compression framework adaptable to various agent memory systems. Specifically, MemForest leverages both global semantic similarity and local temporal continuity of memory events to partition the historical memory into a set of event-centric independent units. For each independent unit, the framework constructs a maximum spanning tree structure, referred to as an EventTree, and performs progressive merging by iteratively selecting high-weight edges, thereby effectively compressing redundant memory nodes and reducing storage overhead. In addition, we introduce an anchor-guided propagation retrieval mechanism, which retrieves more relevant memory nodes from the temporal neighborhoods of key memory nodes, thereby enabling more accurate memory retrieval. Extensive experiments demonstrate the effectiveness of MemForest. Under the unimodal Mem0 framework, across three benchmarks (LoCoMo, LongMemEval, and PersonaMem), MemForest preserves \textbf{97.1\%} of the original performance while compressing \textbf{50\%} of historical memory, achieving a \textbf{1.89×} retrieval speedup. Under the multimodal M3-Agent framework, across two benchmarks (M3-Bench-robot and M3-Bench-web), MemForest retains \textbf{99.7\%} of the original performance under a \textbf{50\%} compression ratio, while achieving a \textbf{2.24×} retrieval speedup. \textcolor{RoyalBlue}{\textit{Our code is available at https://github.com/Celina-love-sweet/MemForest.}}
\end{abstract}

\section{Introduction}

\begin{figure*}[t]
  \centering  
  \includegraphics[width=\textwidth]{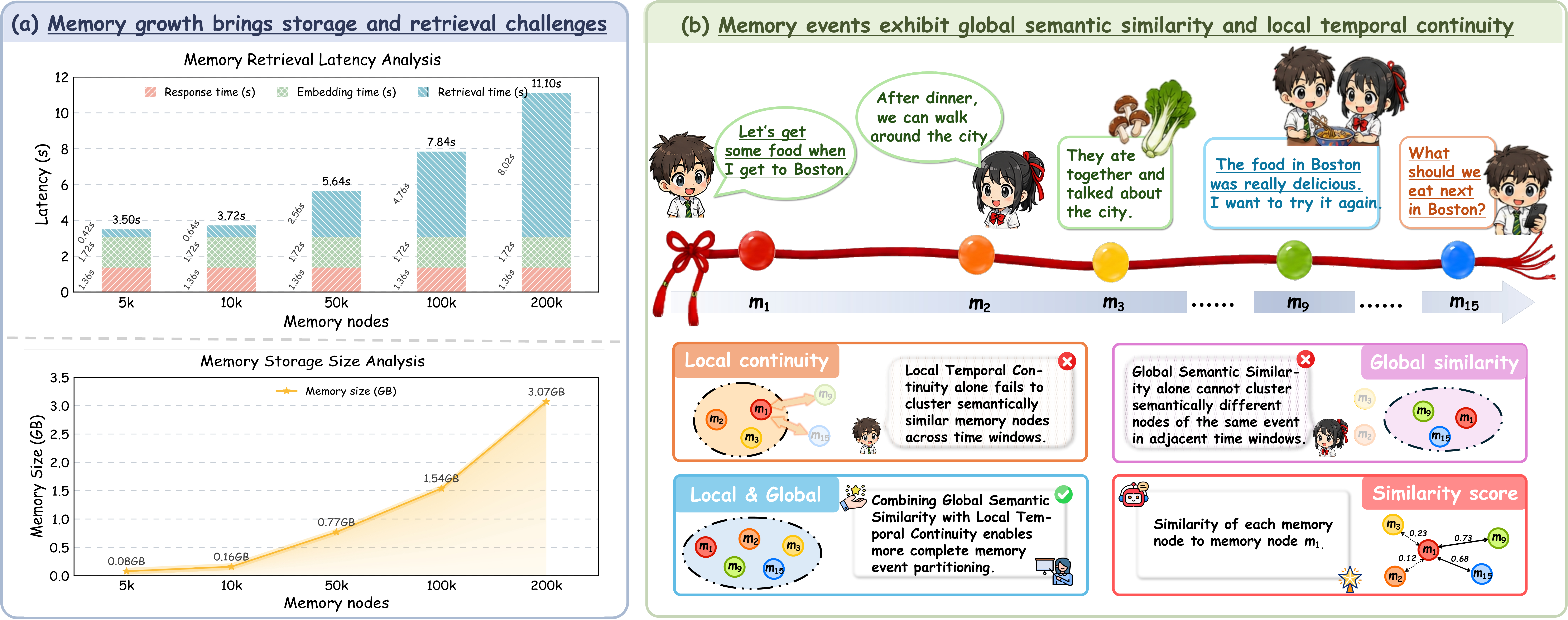}  
  \vspace{-15pt} 
  \caption{\textbf{(a) Challenges brought by the continuous growth of memory.} Each complete user Q\&A involves three time costs: \textit{Embedding time:} converting the query into a vector; \textit{Retrieval time:} searching for corresponding memory nodes via similarity computation; \textit{Response time:} generating the model's answer. \textbf{(b) An example from the LoCoMo benchmark.} This example illustrates that memory events exhibit \textit{global semantic similarity and local temporal continuity}.}
  \label{Fig1}
\vspace{-15pt}
\end{figure*}

In recent years, the application of LLMs has grown rapidly, expanding from dialogue tasks \cite{r48, r47} to a range of problem-solving scenarios \cite{r50, r54, r49, r56, r58, r57, r59, r60, r61, r62}. In these contexts, long-term memory is crucial for maintaining contextual continuity and capturing dependencies across time. To address this, agent memory systems have emerged, storing historical interactions and key information as external memory to enhance a model’s ability to capture long-term dependencies. These systems have shown significant potential in long-term dialogue \cite{r1, r3, r2}, personalized assistants \cite{r6, r4, r5}, and video understanding \cite{r7, r8}. However, as dialogue turns accumulate, memory growth poses challenges to both storage and retrieval efficiency. As shown in Fig.~\ref{Fig1}\textcolor{red}{(a)}, under the Mem0 framework \cite{r1}, expanding memory nodes increases storage pressure and reduces retrieval efficiency. Therefore, exploring efficient memory compression methods has become an important and urgent problem.

However, existing research on memory compression largely focuses on the generation stage \cite{r15, r16}, which cannot fundamentally curb the continuous growth of memory, as stored memories still accumulate over time. Therefore, it is necessary to adopt a post-processing perspective and perform compression on already generated memories to systematically optimize storage efficiency and retrieval performance. Although recent work has begun to explore the feasibility of post-processing memory compression \cite{r38}, this method typically relies on graph-structured memory representations and is restricted to streaming video scenarios, limiting its general applicability. Consequently, designing a general and scalable memory compression framework remains an important and open challenge.

Since stored memories are retrieved as external context, memory compression is closely related to context-length reduction. Recent token compression methods \cite{r9, r34, r12, r33, r32, r11, r10} reduce context via similarity-based merging or diversity-based pruning, mainly for MLLMs. However, unlike redundant visual inputs, textual memories are more semantically rich and information-dense \cite{r14, r13, r10}, making compression challenging; existing methods also overlook memory-specific properties such as event differentiation and temporal relationships. Inspired by the event-level organization of human memory \cite{r45, r46}, we partition historical memory into events and identify two key features for each memory event:

\textbf{(I) Global Semantic Similarity:} Fig.~\ref{Fig1}\textcolor{red}{(b)} suggests that semantically similar memory nodes may recur across distant temporal windows. For example, the similarity between memory node \( m_1\) and the temporally distant memory node \( m_9\) is \textbf{0.73}, and that between node \( m_1\) and node \( m_{15} \) is \textbf{0.68}. This indicates that \textbf{\textit{similar memory nodes within the same event can reappear over long time spans}}.

\textbf{(II) Local Temporal Continuity:} Fig.~\ref{Fig1}\textcolor{red}{(b)} suggests that within adjacent temporal windows, memory nodes may exhibit substantial semantic differences yet still belong to the same event. For example, the similarity between memory node \( m_1\) and node \( m_2\) is only \textbf{0.12}, and that between node \( m_1\) and node \( m_3\) is only \textbf{0.23}, yet all three pertain to "Boston dining plan". This indicates that \textbf{\textit{even when semantic similarity is low, temporally proximate memory nodes may still belong to the same memory event}}.

Consequently, the above analysis indicates that when performing memory compression, it is necessary to consider both global semantic similarity and local temporal continuity. To this end, we propose \textbf{MemForest}, a general agent memory compression framework adaptable to various agent memory systems. Specifically, MemForest leverages the \textbf{EventTree Semantic-Temporal Partitioning} module to combine global semantic similarity with local temporal continuity, thereby partitioning stored historical memory into a set of event-centric EventTrees. For each EventTree, we leverage the \textbf{EventTree Progressive Merging} module to perform iterative processing, progressively compressing its internal representations until a predefined compression ratio is achieved. Furthermore, we introduce an \textbf{Anchor-Guided Propagation Retrieval (AGPR)} mechanism to recover temporal structural information that may be degraded during compression, thereby enabling the retrieval of more relevant memory nodes. In summary, the main contributions are as follows:

\begin{itemize}[leftmargin=1em, itemsep=0pt, topsep=0pt]
    \item \textbf{General Compression Framework.} Based on the above analysis, we propose \textbf{MemForest}, a general agent memory compression framework, which leverages global semantic similarity and local temporal continuity to significantly improve storage and retrieval efficiency.
    \item \textbf{Novel Retrieval Mechanism.} We introduce an \textbf{Anchor-Guided Propagation Retrieval} mechanism to retrieve more relevant memory nodes from the temporal neighborhoods of key memory nodes, thereby enabling more accurate memory retrieval.
    \item \textbf{Excellent Empirical Performance.} Under the Mem0 and M3-Agent frameworks, across several benchmarks, when compressing \textbf{50\%} of the memory, MemForest is able to retain \textbf{97.1\%} and \textbf{99.7\%} of the original performance, respectively, while achieving \textbf{1.89×} and \textbf{2.24×} retrieval speedups.
\end{itemize}

\section{Related Work}

\subsection{Agent Memory Systems}

Agent memory systems \cite{r1, r51, r2, r17, r52, r53} aim to provide agents with long-term storage and retrieval capabilities beyond a single context, thereby supporting persistent interaction and complex task execution. According to the type of input data, they can be broadly categorized into unimodal and multimodal approaches. Unimodal methods \cite{r20, r21, r22} primarily focus on textual data, such as Mem0 \cite{r1}, MemoryOS \cite{r19}, and A-MEM \cite{r18}, which maintain long-term semantic consistency by storing dialogue histories or structured representations. In contrast, multimodal methods \cite{r7, r23, r24} handle more complex inputs, including images, audio, and video. Representative approaches such as MM-MEM \cite{r25}, M3-Agent \cite{r7}, and WorldMM \cite{r8} integrate visual and linguistic information and  further model dynamic scenes such as videos, enabling richer environmental perception and task understanding. However, as memory scales grow, both unimodal and multimodal systems face increasing storage and retrieval costs. This issue is more pronounced in multimodal memory due to greater information redundancy and more complex cross-modal relationships, making efficient memory compression and management a critical challenge.

\subsection{Visual Token Compression}

Visual token compression \cite{r26, r37, r9, r35, r36, r12, r27, r55} aims to reduce the number of tokens while preserving key information as much as possible, thereby lowering computational cost and improving inference efficiency. According to the underlying criteria used for compression, existing methods can be broadly categorized into three types: similarity-based, diversity-based, and attention-based approaches. Similarity-based methods \cite{r26, r28} measure the semantic similarity between tokens and merge highly similar ones to reduce redundant representations. Diversity-based methods \cite{r12, r27} focus on information deduplication by modeling repetitive relationships among tokens, retaining representative tokens while removing redundant information. Attention-based methods \cite{r29, r11, r10} leverage the internal attention mechanism of the model, using attention scores as a measure of token importance to preserve high-weight tokens and discard low-weight ones, enabling adaptive token selection. These methods provide valuable insights for agent memory compression. However, agent memory exhibits unique characteristics such as temporal relationships, and how to effectively leverage these properties for efficient compression remains an open problem.

\section{Methodology}

\subsection{Overview}

\begin{figure*}[t]
  \centering  
  \includegraphics[width=\textwidth]{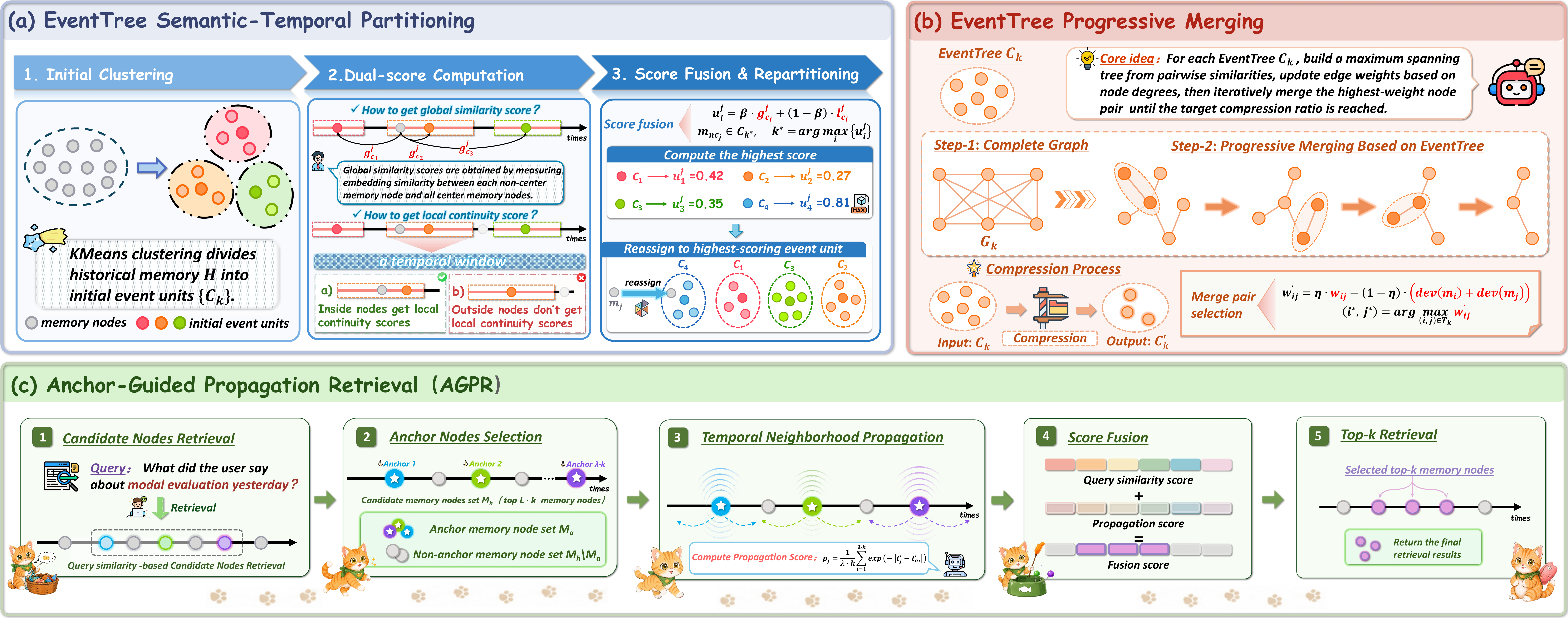}  
  \vspace{-15pt} 
  \caption{\textbf{The overview of our method.} MemForest compresses historical memory through the \textit{EventTree Semantic-Temporal Partitioning} module and the \textit{EventTree Progressive Merging} module, while the \textit{Anchor-Guided Propagation Retrieval} mechanism retrieves more relevant memory nodes within the temporal neighborhoods of key memory nodes, ensuring more accurate retrieval.}
  \label{Fig2}
\vspace{-12pt}
\end{figure*}

The overview of our method is illustrated in Figure~\ref{Fig2}. The MemForest framework consists of two components that achieve efficient compression of historical memory. Meanwhile, the anchor-guided propagation retrieval mechanism enables more accurate retrieval over the compressed memory.

\subsection{EventTree Semantic-Temporal Partitioning}

Given the historical memory \(H\), which consists of a sequence of memory nodes \(m_i = \{c_i, e_i, t_i\}\mid_{i=1}^{N}\), where \(c_i\), \(e_i\), and \(t_i\) denote the content, content embedding, and timestamp of the \(i\)-th memory node, respectively, our goal is to compress the historical memory \(H\) into a compact memory set \(H'\), which consists of a sequence of memory nodes \(\{m'_j\}\mid_{j=1}^{M}\), where \(M < N\).

First, we perform event-unit partitioning of the historical memory \(H\) using the \textit{EventTree Semantic-Temporal Partitioning} module. Specifically, given the initial \(N\) memory nodes \(\{m_i\}\mid_{i=1}^{N}\), we apply the \textit{KMeans clustering algorithm} \cite{r30} to their embedding vectors \(\{e_i\}\mid_{i=1}^{N}\) to partition them into \(N \cdot \alpha\) initial event units, as shown in the following formula:
\begin{equation}
\{C_k\} = \textit{KMeans}(\{e_i\}, \alpha),
\label{eq1}
\end{equation}
where \(\{C_k\}\mid_{k=1}^{N \cdot \alpha}\) denotes the resulting set of event units, and \(\alpha\) is a clustering ratio coefficient used to control the number of event units.

Next, for each event unit, we select the memory node closest to the unit center to form the set of center memory nodes \(H_c\), which consists of a sequence of center nodes \(\{m_{c_i}\}\mid_{i=1}^{N \cdot \alpha}\). Subsequently, for the \(j\)-th non-center memory node \(m_{nc_j} \in H \setminus H_c\), we compute its similarity to all center memory nodes, denoted as \(\{g_{c_i}^j\}\mid_{i=1}^{N \cdot \alpha}\). Each non-center memory node thus obtains \(N \cdot \alpha\) similarity values, which serve as its global similarity scores with respect to each center memory node.

Then, the memory nodes in the historical memory \(H\) are sorted in chronological order, and the positions of all center memory nodes \(\{m_{c_i}\}\mid_{i=1}^{N \cdot \alpha}\) on the timeline are determined. For the \(i\)-th center memory node, a local continuity score is assigned to the non-center memory nodes within its neighboring time window, as defined by the following formula:
\begin{equation}
l_{c_i}^j =
\begin{cases}
1, & \text{if } t_j \in [t_{c_i}-w, t_{c_i}+w] \\
0, & \text{otherwise}
\end{cases},
\label{eq2}
\end{equation}
where \(w\) is the time window coefficient that controls the size of the temporal window around the center memory node, and \(l_{c_i}^j\) represents the local continuity score of the \(j\)-th non-center memory node with respect to the \(i\)-th center memory node.

Subsequently, the global similarity score \(g_{c_i}^j\) of the \(j\)-th non-center memory node with respect to the \(i\)-th center memory node is combined with its local continuity score \(l_{c_i}^j\) to obtain an overall score:
\begin{equation}
u_i^j = \beta \cdot g_{c_i}^j + (1-\beta) \cdot l_{c_i}^j,
\label{eq3}
\end{equation}
where \(\beta\) is a weighting coefficient that controls the relative contribution of the two scores. Each non-center memory node is then reassigned to the event unit with the highest overall score. Taking the \(j\)-th non-center memory node as an example, the assignment can be formulated as:
\begin{equation}
m_{nc_j} \in C_{k^*}, \quad k^* = \arg\max_{i} \{u_i^j\},
\label{eq4}
\end{equation}
resulting in repartitioned event units. For simplicity, these event units are still denoted as \(\{C_k\}\mid_{k=1}^{N \cdot \alpha}\).

\subsection{EventTree Progressive Merging}

For each event unit, we model it as a maximum spanning tree and perform progressive merging via the \textit{EventTree Progressive Merging} module, thereby representing each event unit as an EventTree.

Specifically, for the \(k\)-th EventTree, we denote it as \(C_k = \{m_i^{(k)}\}\mid_{i=1}^{n_k}\), where \(m_i^{(k)}\) denotes the \(i\)-th memory node and \(n_k\) is the number of memory nodes in this EventTree. We first construct a fully connected undirected graph \(G_k = (V_k, E_k)\), where the node set \(V_k = C_k\) and the edge set \(E_k\) consists of all node pairs. For any two nodes \(m_i^{(k)}\) and \(m_j^{(k)}\), the edge weight is defined as the cosine similarity between their embeddings:
\begin{equation}
w_{ij} = \frac{e_i^{(k)} \cdot e_j^{(k)}}{\|e_i^{(k)}\| \cdot \|e_j^{(k)}\|}.
\label{eq5}
\end{equation}
We then construct a maximum spanning tree on \(G_k\) using \textit{Kruskal algorithm} \cite{r31}. Specifically, all edges are first sorted in descending order according to their weights \(w_{ij}\). A union-find structure is used to iteratively add edges with the highest weights while avoiding cycles. This process continues until \(n_k - 1\) edges are selected, resulting in the maximum spanning tree \(T_k\).

After obtaining the maximum spanning tree \(T_k\), we update the edge weights by taking into account the degrees of the two nodes connected by each edge, as follows:
\begin{equation}
w'_{ij} = \eta \cdot w_{ij} - (1-\eta) \cdot \big(\text{deg}(m_i) + \text{deg}(m_j)\big),
\label{eq6}
\end{equation}

where \(\text{deg}(\cdot)\) represents the degree of a node, and \(\eta\) is a weighting coefficient used to balance the two scores. In the maximum spanning tree, nodes with higher degrees are usually central or hub nodes of the event, carrying more core information, and therefore should not be merged prematurely.

We then select the edge with the highest weight:
\begin{equation}
(i^*, j^*) = \arg\max_{(i,j)\in T_k} w'_{ij},
\label{eq7}
\end{equation}
and treat the corresponding nodes \(m_{i^*}^{(k)} = \{c_{i^*}, e_{i^*}, t_{i^*}\}\) and \(m_{j^*}^{(k)} = \{c_{j^*}, e_{j^*}, t_{j^*}\}\) as the optimal merge pair at the current step.

Subsequently, we merge the selected node pair by leveraging a large language model to generate fused content \(c'_l\), re-encoding it to obtain the embedding \(e'_l\), and updating the timestamp to \(t'_l\), resulting in a merged memory node \(m'_l = \{c'_l,\; e'_l,\; t'_l\}\). The merge process can be formally expressed as:
\begin{equation}
c'_l = \mathcal{F}(c_{i^*}, c_{j^*}), \quad
e'_l = \mathcal{E}(c'_l), \quad
t'_l = \max(t_{i^*}, t_{j^*}),
\label{eq8}
\end{equation}
where \(\mathcal{F}(\cdot)\) and \(\mathcal{E}(\cdot)\) denote an external large language model and an embedding model, respectively. The newly generated node \(m'_l\) then replaces the original nodes \(m_{i^*}^{(k)}\) and \(m_{j^*}^{(k)}\).

This process is iteratively applied to the updated node set until a predefined compression ratio is reached. Finally, all the compressed EventTrees \(\{C_k'\}\mid_{k=1}^{N \cdot \alpha}\) collectively form the compact memory set \(H'\), completing the overall compression process.

\subsection{Anchor-Guided Propagation Retrieval}

To enhance the temporal modeling capability of the compressed memory set \(H'\), we propose an \textit{Anchor-Guided Propagation Retrieval} mechanism, which retrieves more relevant memory nodes from the temporal neighborhoods of key memory nodes to supplement nearby information.

Specifically, we first select the top \(L \cdot k\) memory nodes with the highest similarity to the query, forming a candidate set \(M_h\), along with their corresponding similarity scores \(\{s_i\}\mid_{i=1}^{L \cdot k}\), where \(L\) is a scaling coefficient that determines the number of retrieved candidate memory nodes. Then, we further select the top \(\lambda \cdot k\) most similar nodes as anchor memory nodes, denoted as \(M_a = \{m'_{a_i} \}\mid_{i=1}^{\lambda \cdot k}\).

Next, all memory nodes in \(M_h\) are sorted in chronological order, and the positions of anchor memory nodes on the timeline are identified. For the \(j\)-th memory node \( m'_j \in M_h \), we compute its propagation score with respect to all anchor memory nodes as:
\begin{equation}
p_j = \frac{1}{\lambda \cdot k} \sum_{i=1}^{\lambda \cdot k} \exp\left(-\left| t'_j - t'_{a_i} \right|\right).
\label{eq9}
\end{equation}
Finally, we combine the precomputed query similarity and the propagation score to obtain the final fusion score for the \(j\)-th memory node:
\begin{equation}
v_j = \gamma \cdot s_j + (1 - \gamma) \cdot p_j,
\label{eq10}
\end{equation}
where \(\gamma\) is a weighting coefficient that balances the two scores. We rank all memory nodes in \(M_h\) based on \(v_j\) and select the top \(k\) nodes as the final retrieval results.

\section{Theoretical Analysis}
\label{section4}

We analyze from a retrieval perspective \textit{why similar memory nodes should be preferentially merged.}

Given a query \( q \), consider two memory nodes \( m_i \) and \( m_j \) with embeddings \( e_i \) and \( e_j \), respectively. Their similarities to the query are defined as \( s_i = q^\top e_i \) and \( s_j = q^\top e_j \), and the similarity between the two nodes is \( \rho_{ij} = e_i^\top e_j \). We assume all embeddings are \( \ell_2 \)-normalized, i.e., \( \|q\| = \|e_i\| = \|e_j\| = 1 \). Without loss of generality, we assume \( s_i \ge s_j \), indicating that \( m_i \) is more relevant to the query \( q \).

We represent the merged memory embedding as a normalized interpolation of the two embeddings:
\begin{equation}
e_l = \frac{\lambda e_i + (1-\lambda)e_j}{\|\lambda e_i + (1-\lambda)e_j\|},
\label{eq11}
\end{equation}
where \( \lambda \in (0,1) \) is a weighting coefficient that controls the interpolation between \( e_i \) and \( e_j \). 

The similarity between the merged memory node \( m_l \) and the query \( q \) is given by:
\begin{equation}
s_l = \frac{\lambda s_i + (1-\lambda)s_j}{\|\lambda e_i + (1-\lambda)e_j\|}.
\label{eq12}
\end{equation}
By the \textit{Triangle inequality}, we have:
\begin{equation}
s_l \ge \lambda s_i + (1-\lambda)s_j.
\label{eq13}
\end{equation}
Furthermore, by the \textit{Cauchy--Schwarz inequality}, it can be shown that:
\begin{equation}
s_j \ge s_i - \sqrt{2 - 2\rho_{ij}},
\label{eq14}
\end{equation}
which leads to:
\begin{equation}
s_l \ge s_i - (1-\lambda)\sqrt{2 - 2\rho_{ij}}.
\label{eq15}
\end{equation}
The derivation of Eqs.~\ref{eq13}-\ref{eq15} is provided in Appendix~\ref{appendixb.1}. From Eq.~\ref{eq15}, we observe that as \( \rho_{ij} \) increases, the lower bound monotonically increases, indicating that merging highly similar memory nodes preserves higher query similarity. In contrast, when \( \rho_{ij} \) is small, the lower bound decreases, implying larger semantic deviation after merging.

During top \( k \) retrieval, such deviation may prevent important memory nodes from being correctly retrieved. Therefore, it is preferable to prioritize merging highly similar memory nodes.

Furthermore, for the theoretical analysis of MemForest, please refer to Appendix~\ref{appendixb.2}.

\section{Experiments}

\subsection{Experimental Setup}

\paragraph{Baselines.}

For unimodal and multimodal memory systems, we select Mem0 \cite{r1} and M3-Agent \cite{r7} as representative methods, respectively. For the baselines, we categorize them into two groups: pruning-based methods and merging-based methods. Specifically, pruning-based methods include Random Pruning, KMeans \cite{r30}, DART \cite{r12}, and StreamMeCo \cite{r38}; merging-based methods include Random Merging and ToMe \cite{r26}. More details on these methods are provided in Appendix~\ref{appendixa.1}.

\paragraph{Datasets.}

For Mem0 \cite{r1}, we evaluate on three representative benchmarks in the agent memory domain, including LoCoMo \cite{r39}, LongMemEval \cite{r40}, and PersonaMem \cite{r41}. For M3-Agent \cite{r7}, we adopt two representative benchmarks from the streaming video domain, including M3-Bench-robot and M3-Bench-web \cite{r7}. More details on these benchmarks are provided in Appendix~\ref{appendixa.2}.

\paragraph{Implementation Details.}

All experiments are conducted on two NVIDIA A100 (80GB) GPUs. The key hyperparameters are set as follows: \(\alpha = 0.05\), \(\beta = 0.8\), \(w = 5\), \(\eta = 0.99\), \(\gamma = 0.9\), \(L = 4\), and \(\lambda = 0.2\). We use GPT-4o-mini for memory node merging. In addition, all other settings follow the original papers \cite{r1, r7}. All experimental results are averaged over three runs. More details are provided in Appendix~\ref{appendixa.3}.

\begin{table*}[t]
\fontsize{8}{13}\selectfont
\renewcommand{\arraystretch}{0.8}
\setlength{\tabcolsep}{2.5pt}
\centering
\caption{\textbf{Performance comparison of different baselines on Mem0 under varying compression ratios.} {\color{blue!30}Blue} denotes pruning methods, while {\color{red!30}Red} denotes merging methods. \textbf{\%} represents the performance retention rate. For each ratio, the top two results are highlighted in \textbf{bold black font}.}
\vspace{3pt} 
\begin{tabular}{lccccc|ccccccc|c|c|c}
  \toprule[1.5pt]
  \textbf{Dataset}
       & \multicolumn{5}{c|}{\textbf{LoCoMo}}
       & \multicolumn{7}{c|}{\textbf{LongMemEval}}
       & \multirow{2}{*}[-0.8ex]{\shortstack[c]{\textbf{Persona}\\\textbf{-Mem}}}
       & \multirow{2}{*}[-0.8ex]{\textbf{Avg.}}
       & \multirow{2}{*}[-0.8ex]{\textbf{\%}} \\
  \cmidrule(lr){2-6}\cmidrule(lr){7-13}
  \textbf{Method}
       & SH & MH & TR & OD & \textbf{All}
       & SU & SA & SP & MS & KU & TR & \textbf{All}
       &  &  &  \\
  \midrule

  Mem0  &66.0&56.0&53.9&42.7&60.2&90.0&21.4&23.3&54.9&79.5&44.4&55.2&67.9&61.1& 100.0\% \\
  \hspace{7pt}+ AGPR
        &66.2&56.4&53.9&40.6&60.3&90.0&21.4&23.3&54.1&78.2&43.6&54.6&68.6&61.2& 100.2\% \\

  \rowcolor{gray!15}
  & \multicolumn{13}{c}{\emph{Historical memory compression ratio ({\color{green!40!black}$\downarrow 30\%$})}}
  & \multicolumn{2}{c}{~} \\

  \rowcolor{blue!3}
  \hspace{7pt}+ Random Pruning
        &55.3&49.7&43.3&39.6&50.8&82.9&17.9&26.7&51.9&73.1&39.1&50.8&64.5&55.4& 90.7\% \\
  \rowcolor{blue!3}
  \hspace{7pt}+ KMeans
        &61.1&54.3&52.7&40.6&56.8&84.3&17.9&20.0&51.9&75.6&41.4&52.0&65.7&58.2& 95.3\% \\
  \rowcolor{blue!3}
  \hspace{7pt}+ DART
        &64.1&53.9&48.0&40.6&57.4&82.9&19.0&20.0&48.9&74.4&34.6&48.8&\textbf{66.6}&57.6& 94.3\% \\
  \rowcolor{red!3}
  \hspace{7pt}+ Random Merging
        &60.9&52.1&51.1&43.8&56.2&91.4&21.4&20.2&46.6&74.4&41.4&50.8&65.2&57.4& 93.9\% \\
  \rowcolor{red!3}
  \hspace{7pt}+ ToMe
        &63.1&51.1&50.8&44.8&57.2&85.7&21.4&23.3&48.1&71.8&42.9&51.2&66.4&58.3& 95.4\% \\
  \rowcolor{red!3}
  \hspace{7pt}+ MemForest
        &63.7&54.6&52.0&45.8&\textbf{58.5}&90.0&19.6&20.0&49.8&76.9&49.6&\textbf{54.2}&66.4&\textbf{59.7}& \textbf{97.7\%} \\
  \rowcolor{red!3}
  \hspace{7pt}+ MemForest + AGPR
        &64.6&54.6&52.3&44.8&\textbf{59.0}&90.0&19.6&20.0&52.6&78.8&43.6&\textbf{53.8}&\textbf{67.6}&\textbf{60.1}& \textbf{98.4\%} \\

  \midrule

  \rowcolor{gray!15}
  & \multicolumn{13}{c}{\emph{Historical memory compression ratio ({\color{green!40!black}$\downarrow 50\%$})}}
  & \multicolumn{2}{c}{~} \\

  \rowcolor{blue!3}
  \hspace{7pt}+ Random Pruning
        &44.4&45.4&37.1&44.8&43.1&75.7&17.9&26.7&32.3&65.4&39.8&43.6&64.3&50.3& 82.3\% \\
  \rowcolor{blue!3}
  \hspace{7pt}+ KMeans
        &51.6&43.6&39.6&37.5&46.8&77.1&19.6&23.3&39.1&74.4&39.8&47.0&65.2&53.0& 86.7\% \\
  \rowcolor{blue!3}
  \hspace{7pt}+ DART
        &57.4&49.6&41.7&43.8&51.9&67.1&12.5&23.3&42.9&60.3&31.6&41.4&64.3&52.5& 85.9\% \\
  \rowcolor{red!3}
  \hspace{7pt}+ Random Merging
        &56.0&48.2&46.4&37.5&51.4&85.7&23.2&20.0&42.9&70.5&39.1&48.6&64.5&54.8& 89.7\% \\
  \rowcolor{red!3}
  \hspace{7pt}+ ToMe
        &60.2&50.4&47.4&43.8&54.7&85.7&21.4&23.3&48.1&71.8&41.4&50.8&65.9&57.1& 93.5\% \\
  \rowcolor{red!3}
  \hspace{7pt}+ MemForest
        &62.1&56.7&54.8&43.8&\textbf{58.4}&90.0&19.6&20.0&48.1&74.4&42.9&\textbf{51.8}&\textbf{67.7}&\textbf{59.3}& \textbf{97.1\%} \\
  \rowcolor{red!3}
  \hspace{7pt}+ MemForest + AGPR
        &63.7&57.1&52.3&45.8&\textbf{59.0}&90.0&21.4&23.3&47.4&76.9&45.1&\textbf{53.0}&\textbf{68.1}&\textbf{60.0}& \textbf{98.3\%} \\

  \midrule

  \rowcolor{gray!15}
  & \multicolumn{13}{c}{\emph{Historical memory compression ratio ({\color{green!40!black}$\downarrow 70\%$})}}
  & \multicolumn{2}{c}{~} \\
  
  \rowcolor{blue!3}
  \hspace{7pt}+ Random Pruning
        &35.0&36.5&25.9&31.3&33.1&47.1&7.1&30.0&16.5&42.3&31.6&28.6&63.7&41.8& 68.4\% \\
  \rowcolor{blue!3}
  \hspace{7pt}+ KMeans
        &37.9&42.6&27.4&39.6&36.7&61.4&8.9&30.0&21.8&59.0&31.6&34.8&63.8&45.1& 73.8\% \\
  \rowcolor{blue!3}
  \hspace{7pt}+ DART
        &46.4&44.3&36.1&41.7&43.6&44.3&10.7&16.7&21.0&43.6&27.8&30.6&61.8&45.3& 74.1\% \\
  \rowcolor{red!3}
  \hspace{7pt}+ Random Merging
        &54.0&47.2&47.0&43.8&50.7&80.0&21.4&30.0&41.4&67.9&34.6&46.2&64.0&53.6& 87.7\% \\
  \rowcolor{red!3}
  \hspace{7pt}+ ToMe
        &56.6&51.8&47.7&38.5&52.7&84.3&21.4&20.0&45.1&70.5&42.1&49.6&\textbf{65.7}&56.0& 91.7\% \\
  \rowcolor{red!3}
  \hspace{7pt}+ MemForest
        &58.1&56.4&45.2&45.8&\textbf{54.4}&84.3&25.0&23.3&52.6&74.4&39.8&\textbf{52.2}&64.4&\textbf{57.0}& \textbf{93.3\%} \\
  \rowcolor{red!3}
  \hspace{7pt}+ MemForest + AGPR
        &57.7&55.0&45.6&45.8&\textbf{53.9}&85.7&26.8&26.7&51.1&74.4&39.8&\textbf{52.4}&\textbf{65.5}&\textbf{57.3}& \textbf{93.8\%} \\

  \bottomrule[1.5pt]
\end{tabular}
\label{table1}
\vspace{-12pt} 
\end{table*}

\begin{table*}[t]
\fontsize{8}{13}\selectfont
\renewcommand{\arraystretch}{0.8}
\setlength{\tabcolsep}{3.6pt}
\centering
\caption{\textbf{Performance comparison of different baselines on M3-Agent under varying compression ratios.} {\color{blue!30}Blue} denotes pruning methods, while {\color{red!30}Red} denotes merging methods. \textbf{\%} represents the performance retention rate. For each ratio, the top two results are highlighted in \textbf{bold black font}.}
\vspace{3pt} 
\begin{tabular}{lcccccc|cccccc|c|c}
  \toprule[1.5pt]
  \textbf{Dataset}
       & \multicolumn{6}{c|}{\textbf{M3-Bench-robot}}
       & \multicolumn{6}{c|}{\textbf{M3-Bench-web}}
       & \multirow{2}{*}[-0.8ex]{\textbf{Avg.}}
       & \multirow{2}{*}[-0.8ex]{\textbf{\%}} \\
  \cmidrule(lr){2-7}\cmidrule(lr){8-13}
  \textbf{Method}
       & ME & MH & CM & PU & GK & \textbf{All}
       & ME & MH & CM & PU & GK & \textbf{All}
       &  &  \\
  \midrule

  M3-Agent  &30.9&29.4&29.6&41.1&22.6&30.3&44.9&25.6&44.8&58.6&53.7&47.9&39.1 & 100.0\% \\
  \hspace{7pt}+ AGPR
            &37.2&38.8&34.2&47.3&29.7&37.2&49.2&35.9&49.8&64.5&61.2&54.8&46.0 & 117.6\% \\

  \rowcolor{gray!15}
  & \multicolumn{12}{c}{\emph{Historical memory compression ratio ({\color{green!40!black}$\downarrow 30\%$})}}
  & \multicolumn{2}{c}{~} \\

  \rowcolor{blue!3}
  \hspace{7pt}+ Random Pruning
            &31.4&27.1&29.8&42.5&18.3&29.1&40.8&23.9&41.7&56.0&49.1&44.9&37.0 & 94.6\% \\
  \rowcolor{blue!3}
  \hspace{7pt}+ KMeans
            &31.6&32.9&29.6&42.2&19.3&30.2&41.0&24.1&45.0&56.1&49.8&45.8&38.0 & 97.2\% \\
  \rowcolor{blue!3}
  \hspace{7pt}+ DART
            &31.8&29.4&32.6&43.8&19.3&29.6&40.9&24.9&43.4&58.2&49.6&46.1&37.8 & 96.7\% \\
  \rowcolor{blue!3}
  \hspace{7pt}+ StreamMeCo
            &32.9&30.6&30.9&42.5&19.6&30.7&41.3&26.0&44.3&58.3&50.5&47.0&38.9 & 99.5\% \\
  \rowcolor{red!3}
  \hspace{7pt}+ Random Merging
            &31.7&35.3&30.3&40.9&19.9&30.0&40.4&23.2&42.9&55.9&50.1&45.1&37.6 & 96.2\% \\
  \rowcolor{red!3}
  \hspace{7pt}+ ToMe
            &30.8&25.9&27.7&40.5&21.7&30.4&42.4&27.6&41.7&57.5&52.0&47.1&38.8 & 99.2\% \\
  \rowcolor{red!3}
  \hspace{7pt}+ MemForest
            &32.3&27.1&33.6&43.8&22.6&\textbf{32.0}&43.1&26.3&44.6&57.9&52.2&\textbf{47.2}&\textbf{39.6} & \textbf{101.3}\% \\
  \rowcolor{red!3}
  \hspace{7pt}+ MemForest + AGPR
            &39.4&37.6&36.8&50.5&29.4&\textbf{39.0}&50.9&39.2&52.4&64.0&63.0&\textbf{56.0}&\textbf{47.5} & \textbf{121.5}\% \\

  \midrule

  \rowcolor{gray!15}
  & \multicolumn{12}{c}{\emph{Historical memory compression ratio ({\color{green!40!black}$\downarrow 50\%$})}}
  & \multicolumn{2}{c}{~} \\

  \rowcolor{blue!3}
  \hspace{7pt}+ Random Pruning
            &28.4&25.9&27.3&38.7&19.6&28.1&35.6&23.0&34.9&52.3&44.2&41.2&34.7 & 88.7\% \\
  \rowcolor{blue!3}
  \hspace{7pt}+ KMeans
            &29.1&30.6&28.4&41.4&20.8&29.2&41.1&21.7&42.5&55.0&47.4&43.9&36.6 & 93.6\% \\
  \rowcolor{blue!3}
  \hspace{7pt}+ DART
            &28.9&25.9&29.4&39.1&22.0&29.1&39.9&23.6&39.4&52.9&47.2&43.1&36.1 & 92.3\% \\
  \rowcolor{blue!3}
  \hspace{7pt}+ StreamMeCo
            &32.3&28.2&29.8&41.4&21.1&30.6&39.7&25.2&38.9&58.4&47.3&44.7&37.7 & 96.2\% \\
  \rowcolor{red!3}
  \hspace{7pt}+ Random Merging
            &33.0&32.9&31.9&44.3&19.6&30.8&41.8&21.0&40.1&58.0&46.3&44.6&37.7 & 96.4\% \\
  \rowcolor{red!3}
  \hspace{7pt}+ ToMe
            &30.5&32.9&29.6&42.3&19.9&30.9&42.7&24.7&41.7&54.3&53.1&46.1&38.5 & 98.5\% \\
  \rowcolor{red!3}
  \hspace{7pt}+ MemForest
            &31.6&31.8&31.1&45.1&21.4&\textbf{31.6}&42.6&23.9&43.4&57.1&51.5&\textbf{46.5}&\textbf{39.0} & \textbf{99.7}\% \\
  \rowcolor{red!3}
  \hspace{7pt}+ MemForest + AGPR
            &40.5&38.8&36.1&48.2&31.8&\textbf{38.3}&52.8&34.4&50.0&65.0&63.8&\textbf{56.1}&\textbf{47.2} & \textbf{120.7}\% \\

  \midrule

  \rowcolor{gray!15}
  & \multicolumn{12}{c}{\emph{Historical memory compression ratio ({\color{green!40!black}$\downarrow 70\%$})}}
  & \multicolumn{2}{c}{~} \\

  \rowcolor{blue!3}
  \hspace{7pt}+ Random Pruning
            &27.3&25.9&26.7&39.2&20.8&27.7&35.0&19.9&34.9&48.7&39.5&37.3&32.5 & 83.1\% \\
  \rowcolor{blue!3}
  \hspace{7pt}+ KMeans
            &28.7&28.2&27.3&42.7&19.3&28.9&35.5&20.1&34.9&39.7&42.5&39.1&34.0 & 87.0\% \\
  \rowcolor{blue!3}
  \hspace{7pt}+ DART
            &27.8&25.9&25.4&40.0&16.2&27.2&35.4&19.5&36.6&48.2&45.0&39.2&33.2 & 84.9\% \\
  \rowcolor{blue!3}
  \hspace{7pt}+ StreamMeCo
            &29.1&30.6&26.9&40.3&19.9&28.4&37.0&21.7&36.1&53.0&46.3&41.9&35.1 & 89.8\% \\
  \rowcolor{red!3}
  \hspace{7pt}+ Random Merging
            &30.6&31.8&28.8&45.1&19.0&30.4&41.1&21.9&39.6&58.6&41.5&43.2&36.8 & 94.1\% \\
  \rowcolor{red!3}
  \hspace{7pt}+ ToMe
            &31.7&27.1&28.6&43.1&20.2&30.6&42.9&23.6&43.6&56.5&49.8&45.7&38.2 & 97.7\% \\
  \rowcolor{red!3}
  \hspace{7pt}+ MemForest
            &31.2&32.9&32.4&43.6&17.4&\textbf{30.7}&44.8&23.9&38.9&57.3&50.4&\textbf{46.3}&\textbf{38.5} & \textbf{98.5}\% \\
  \rowcolor{red!3}
  \hspace{7pt}+ MemForest + AGPR
            &38.5&38.8&35.7&48.2&29.7&\textbf{37.5}&52.8&34.1&50.0&64.1&64.3&\textbf{55.8}&\textbf{46.7} & \textbf{119.4}\% \\

  \bottomrule[1.5pt]
\end{tabular}
\label{table2}
\vspace{-5pt} 
\end{table*}

\subsection{Main Results}

\textbf{Performance under the Mem0 Framework.} Table~\ref{table1} reports the performance of various baselines under the Mem0 framework. The experimental results can be summarized in three points: \textbf{(i) Outstanding performance.} Without incorporating the AGPR mechanism, MemForest is able to retain \textbf{97.1\%} of the original performance across three representative benchmarks even when \textbf{50\%} of historical memory is compressed, significantly outperforming various pruning-based and merging-based baselines. \textbf{(ii) Historical information preservation.} At a low compression ratio (30\%), pruning-based and merging-based methods perform comparably. However, when the compression ratio exceeds \textbf{50\%}, pruning-based methods suffer a noticeable performance drop, whereas merging-based methods still maintain relatively strong performance and preserve more historical information. \textbf{(iii) Dataset-specific differences.} On the PersonaMem benchmark, all methods exhibit relatively minor performance degradation under different compression ratios. We attribute this to the multiple-choice nature of the questions, which requires only limited memory for decision-making.

\textbf{Performance under the M3-Agent Framework.} Table~\ref{table2} presents the performance of various baselines under the M3-Agent framework. The experimental results can be summarized as follows: \textbf{(i) Outstanding performance.} Without incorporating the AGPR mechanism, MemForest is able to retain \textbf{99.7\%} of the original performance across two representative benchmarks even when \textbf{50\%} of historical memory is compressed, significantly outperforming various pruning-based and merging-based baselines. \textbf{(ii) Historical information preservation.} When the compression ratio exceeds \textbf{50\%}, even the memory pruning framework StreamMeCo, which is specifically designed for M3-Agent, performs worse than random compression, indicating that merging-based methods better preserve historical information. \textbf{(iii) Higher redundancy.} Compared with Mem0, the performance degradation of M3-Agent is considerably smaller, indicating that multimodal memory with dense visual inputs has higher redundancy than unimodal text memory, offering greater potential for compression.

\textbf{Performance of the AGPR Mechanism.} Tables~\ref{table1}–\ref{table2} illustrate the performance of the AGPR mechanism across the two memory frameworks. The experimental results can be summarized in three points: \textbf{(i) Outstanding performance.} The AGPR mechanism consistently improves performance across different compression ratios in both memory frameworks. \textbf{(ii) Framework-dependent performance gains.} Under the Mem0 framework, the performance improvement brought by AGPR is relatively limited and noticeably smaller than that in M3-Agent. This difference mainly stems from differences in the original retrieval mechanism and the nature of the input data. The original retrieval mechanism of M3-Agent is relatively weak; in addition, Mem0 uses dense textual input with low redundancy, whereas M3-Agent contains multimodal input with substantial redundancy, allowing AGPR to retrieve key memory nodes more accurately. \textbf{(iii) Information loss compensation.} Performance gains are more pronounced after compressing historical memory than without compression. For example, when \textbf{50\%} of historical memory is compressed, performance improvements in Mem0 and M3-Agent are \textbf{1.2\%} and \textbf{21.0\%}, respectively, compared to \textbf{0.2\%} and \textbf{17.6\%} without compression. This demonstrates that AGPR can effectively compensate for information loss by enriching the contextual information around key memory nodes.

\subsection{Ablation Study}
\label{section5.3}

\subsubsection{MemForest Ablation Study.}

\begin{table}[t]
\centering
\begin{minipage}{0.48\linewidth}
\centering
\caption{\textbf{Impact of different EventTree partitioning strategies at 50\% compression rate.}}
\vspace{-3pt} 
\fontsize{7}{10}\selectfont
\setlength{\tabcolsep}{1.2pt}
\setlength{\abovecaptionskip}{4pt}
\setlength{\belowcaptionskip}{-10pt}
\begin{tabular}{l|ccc|c|c}
\toprule[1.2pt]
\textbf{Method} & \textbf{LoCoMo} & \textbf{LongMemEval} & \textbf{M3-Bench-web} & \textbf{Avg.} & \textbf{\%} \\
\midrule
\rowcolor{gray!7}
Baseline & 60.2 & 55.2 & 47.9 & 54.4 & 100.0\% \\
w/o \(g_{c_i}^j\) & 54.5 & 49.2 & 45.6 & 49.8 & 91.5\% \\
w/o \(l_{c_i}^j\) & 56.4 & 50.2 & 46.2 & 50.9 & 93.6\% \\
\rowcolor{green!5}
\textbf{Ours} & \textbf{58.4} & \textbf{51.8} & \textbf{46.5} & \textbf{52.2} & \textbf{96.0\%} \\
\bottomrule[1.2pt]
\end{tabular}
\label{table3}
\end{minipage}
\hspace{0.01\linewidth}
\begin{minipage}{0.48\linewidth}
\centering
\caption{\textbf{Impact of different memory node merging strategies at 50\% compression rate.}}
\vspace{-3pt}
\fontsize{7}{10}\selectfont
\setlength{\tabcolsep}{1.2pt}
\setlength{\abovecaptionskip}{4pt}
\setlength{\belowcaptionskip}{-10pt}
\begin{tabular}{l|ccc|c|c}
\toprule[1.2pt]
\textbf{Method} & \textbf{LoCoMo} & \textbf{LongMemEval} & \textbf{M3-Bench-web} & \textbf{Avg.} & \textbf{\%} \\
\midrule
\rowcolor{gray!7}
Baseline & 60.2 & 55.2 & 47.9 & 54.4 & 100.0\% \\
Random & 54.0 & 50.0 & 45.7 & 49.9 & 91.7\% \\
Minimum & 51.2 & 48.8 & 45.2 & 48.4 & 89.0\% \\
\rowcolor{green!5}
\textbf{Ours} & \textbf{58.4} & \textbf{51.8} & \textbf{46.5} & \textbf{52.2} & \textbf{96.0\%} \\
\bottomrule[1.2pt]
\end{tabular}
\label{table4}
\end{minipage}
\end{table}

\begin{figure*}[t]
\vspace{-10pt}
  \centering  
  \includegraphics[width=\textwidth]{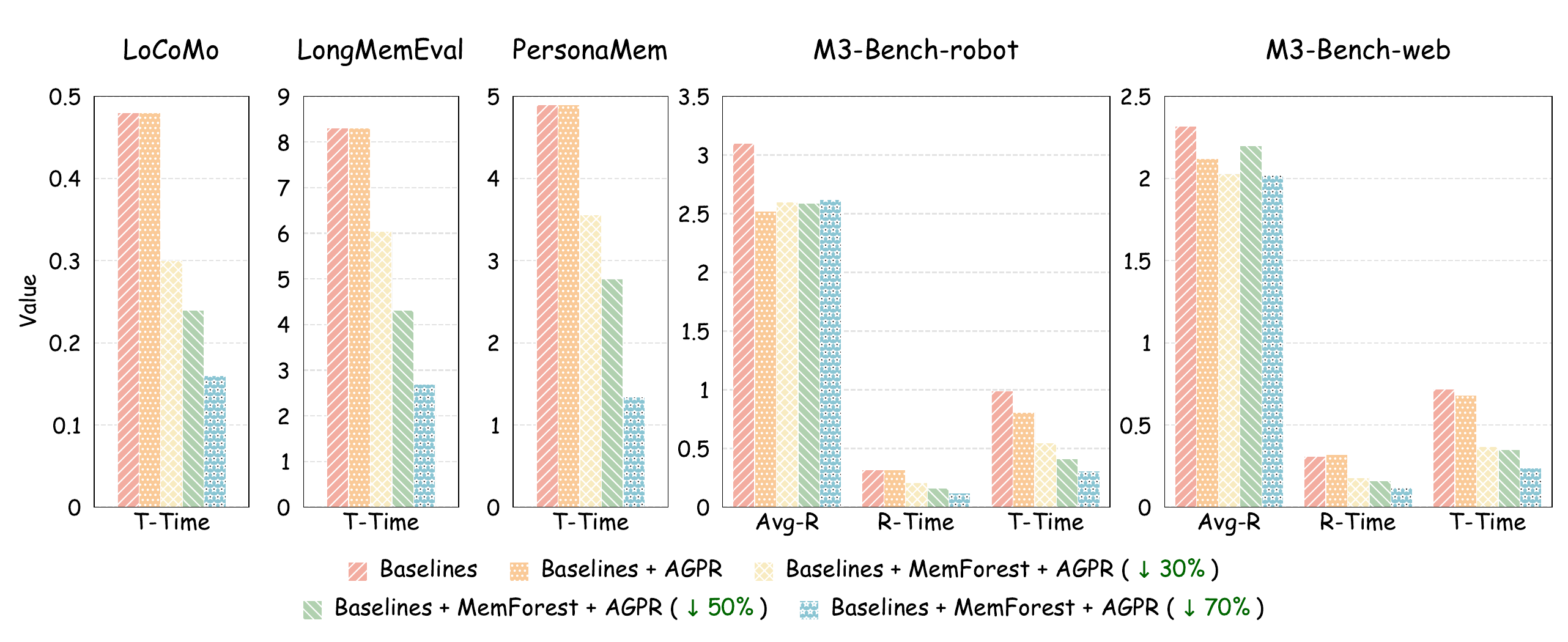}  
  \vspace{-20pt} 
  \caption{\textbf{Time Efficiency Analysis on Mem0 and M3-Agent}. \textit{Avg-R(×)}: average number of retrievals per query; \textit{R-Time(s)}: time cost per retrieval; \textit{T-Time(s)}: total retrieval time per query.}
  \label{Fig3}
  \vspace{-12pt} 
\end{figure*}

\textbf{EventTree Semantic-Temporal Partitioning.} Table~\ref{table3} suggests that, when compressing \textbf{50\%} of historical memory, our method significantly outperforms two variants across three benchmarks: one removes the local continuity score \(l_{c_i}^j\), and the other removes the global similarity score \(g_{c_i}^j\) and replaces it with a sliding window for EventTree partitioning. Detailed implementations of these variants are provided in Appendix~\ref{appendixa.3}. This indicates that combining global semantic similarity with local temporal continuity enables more accurate EventTree partitioning.

\textbf{EventTree Progressive Merging.} Table~\ref{table4} suggests that, when compressing \textbf{50\%} of the historical memory, replacing the maximum spanning tree with either random merging or minimum spanning tree merging results in significantly worse performance than our method, thereby validating the theoretical analysis in Section~\ref{section4}.

\subsubsection{Time Efficiency Analysis.}

Figure~\ref{Fig3} suggests our proposed method achieves high time efficiency, and detailed results can be found in Appendix~\ref{appendixd.7}, from which we draw three key observations: \textbf{(i) Minimal time overhead.} Since the AGPR mechanism avoids complex embedding similarity computations, it introduces negligible additional time overhead. \textbf{(ii) Significant speedup.} Under the Mem0 framework, when compressing \textbf{50\%} of historical memory across three benchmarks, our method achieves speedups of \textbf{2.00$\times$}, \textbf{1.92$\times$}, and \textbf{1.76$\times$}, respectively, with an average speedup of \textbf{1.89$\times$}. \textbf{(iii) More accurate retrieval.} M3-Agent typically requires multiple rounds of memory retrieval for a single query to obtain sufficient information, whereas our method significantly reduces the number of retrieval iterations, indicating more accurate retrieval of critical memory nodes. As a result, our method achieves speedups of \textbf{2.41$\times$} and \textbf{2.06$\times$} on two benchmarks, respectively, with an average speedup of \textbf{2.24$\times$}.

For more ablation studies, additional evaluation metrics, parameter sensitivity analyses, merge cost analysis, analysis of merging effectiveness across different models, case studies, discussions on future work, etc., please refer to Appendices~\ref{appendixd}-\ref{appendixg}.

\section{Conclusion}

We propose MemForest, a general agent memory compression framework, and introduce an anchor-guided propagation retrieval mechanism to retrieve more relevant neighborhood information. Extensive experiments demonstrate the effectiveness of both components. Under the Mem0 and M3-Agent frameworks, when compressing 50\% of historical memory, MemForest retains 97.1\% and 99.7\% of the original performance, while achieving 1.89$\times$ and 2.24$\times$ retrieval speedups, respectively.

\section*{Acknowledgments}

This paper was partially sponsored by WUYING - Alibaba Cloud.

\small
\bibliographystyle{plain}
\bibliography{Reference}

@article{r1,
  title={Mem0: Building production-ready ai agents with scalable long-term memory},
  author={Chhikara, Prateek and Khant, Dev and Aryan, Saket and Singh, Taranjeet and Yadav, Deshraj},
  journal={arXiv preprint arXiv:2504.19413},
  year={2025}
}

@article{r2,
  title={Memos: A memory os for ai system},
  author={Li, Zhiyu and Xi, Chenyang and Li, Chunyu and Chen, Ding and Chen, Boyu and Song, Shichao and Niu, Simin and Wang, Hanyu and Yang, Jiawei and Tang, Chen and others},
  journal={arXiv preprint arXiv:2507.03724},
  year={2025}
}

@article{r3,
  title={EverMemOS: A Self-Organizing Memory Operating System for Structured Long-Horizon Reasoning},
  author={Hu, Chuanrui and Gao, Xingze and Zhou, Zuyi and Xu, Dannong and Bai, Yi and Li, Xintong and Zhang, Hui and Li, Tong and Zhang, Chong and Bing, Lidong and others},
  journal={arXiv preprint arXiv:2601.02163},
  year={2026}
}

@inproceedings{r4,
  title={Memoria: A scalable agentic memory framework for personalized conversational ai},
  author={Sarin, Samarth and Singh, Lovepreet and Sarmah, Bhaskarjit and Mehta, Dhagash},
  booktitle={2025 5th International Conference on AI-ML-Systems (AIMLSystems)},
  pages={32--39},
  year={2025},
  organization={IEEE}
}

@article{r5,
  title={O-mem: Omni memory system for personalized, long horizon, self-evolving agents},
  author={Wang, Piaohong and Tian, Motong and Li, Jiaxian and Liang, Yuan and Wang, Yuqing and Chen, Qianben and Wang, Tiannan and Lu, Zhicong and Ma, Jiawei and Jiang, Yuchen Eleanor and others},
  journal={arXiv preprint arXiv:2511.13593},
  year={2025}
}

@article{r6,
  title={On memory construction and retrieval for personalized conversational agents},
  author={Pan, Zhuoshi and Wu, Qianhui and Jiang, Huiqiang and Luo, Xufang and Cheng, Hao and Li, Dongsheng and Yang, Yuqing and Lin, Chin-Yew and Zhao, H Vicky and Qiu, Lili and others},
  journal={arXiv preprint arXiv:2502.05589},
  year={2025}
}

@article{r7,
  title={Seeing, listening, remembering, and reasoning: A multimodal agent with long-term memory},
  author={Long, Lin and He, Yichen and Ye, Wentao and Pan, Yiyuan and Lin, Yuan and Li, Hang and Zhao, Junbo and Li, Wei},
  journal={arXiv preprint arXiv:2508.09736},
  year={2025}
}

@article{r8,
  title={Worldmm: Dynamic multimodal memory agent for long video reasoning},
  author={Yeo, Woongyeong and Kim, Kangsan and Yoon, Jaehong and Hwang, Sung Ju},
  journal={arXiv preprint arXiv:2512.02425},
  year={2025}
}

@inproceedings{r9,
  title={An image is worth 1/2 tokens after layer 2: Plug-and-play inference acceleration for large vision-language models},
  author={Chen, Liang and Zhao, Haozhe and Liu, Tianyu and Bai, Shuai and Lin, Junyang and Zhou, Chang and Chang, Baobao},
  booktitle={European Conference on Computer Vision},
  pages={19--35},
  year={2024},
  organization={Springer}
}

@article{r10,
  title={Sparsevlm: Visual token sparsification for efficient vision-language model inference},
  author={Zhang, Yuan and Fan, Chun-Kai and Ma, Junpeng and Zheng, Wenzhao and Huang, Tao and Cheng, Kuan and Gudovskiy, Denis and Okuno, Tomoyuki and Nakata, Yohei and Keutzer, Kurt and others},
  journal={arXiv preprint arXiv:2410.04417},
  year={2024}
}

@article{r11,
  title={[CLS] Attention is All You Need for Training-Free Visual Token Pruning: Make VLM Inference Faster},
  author={Zhang, Qizhe and Cheng, Aosong and Lu, Ming and Zhuo, Zhiyong and Wang, Minqi and Cao, Jiajun and Guo, Shaobo and She, Qi and Zhang, Shanghang},
  journal={arXiv e-prints},
  pages={arXiv--2412},
  year={2024}
}

@inproceedings{r12,
  title={Stop Looking for “Important Tokens” in Multimodal Language Models: Duplication Matters More},
  author={Wen, Zichen and Gao, Yifeng and Wang, Shaobo and Zhang, Junyuan and Zhang, Qintong and Li, Weijia and He, Conghui and Zhang, Linfeng},
  booktitle={Proceedings of the 2025 Conference on Empirical Methods in Natural Language Processing},
  pages={9972--9991},
  year={2025}
}

@inproceedings{r13,
  title={Beyond text-visual attention: Exploiting visual cues for effective token pruning in vlms},
  author={Zhang, Qizhe and Cheng, Aosong and Lu, Ming and Zhang, Renrui and Zhuo, Zhiyong and Cao, Jiajun and Guo, Shaobo and She, Qi and Zhang, Shanghang},
  booktitle={Proceedings of the IEEE/CVF International Conference on Computer Vision},
  pages={20857--20867},
  year={2025}
}

@book{r14,
  title={Vision: A computational investigation into the human representation and processing of visual information},
  author={Marr, David},
  year={2010},
  publisher={MIT press}
}

@article{r15,
  title={Lightmem: Lightweight and efficient memory-augmented generation},
  author={Fang, Jizhan and Deng, Xinle and Xu, Haoming and Jiang, Ziyan and Tang, Yuqi and Xu, Ziwen and Deng, Shumin and Yao, Yunzhi and Wang, Mengru and Qiao, Shuofei and others},
  journal={arXiv preprint arXiv:2510.18866},
  year={2025}
}

@article{r16,
  title={SimpleMem: Efficient Lifelong Memory for LLM Agents},
  author={Liu, Jiaqi and Su, Yaofeng and Xia, Peng and Han, Siwei and Zheng, Zeyu and Xie, Cihang and Ding, Mingyu and Yao, Huaxiu},
  journal={arXiv preprint arXiv:2601.02553},
  year={2026}
}

@article{r17,
  title={General agentic memory via deep research},
  author={Yan, BY and Li, Chaofan and Qian, Hongjin and Lu, Shuqi and Liu, Zheng},
  journal={arXiv preprint arXiv:2511.18423},
  year={2025}
}

@article{r18,
  title={A-mem: Agentic memory for llm agents},
  author={Xu, Wujiang and Liang, Zujie and Mei, Kai and Gao, Hang and Tan, Juntao and Zhang, Yongfeng},
  journal={arXiv preprint arXiv:2502.12110},
  year={2025}
}

@inproceedings{r19,
  title={Memory os of ai agent},
  author={Kang, Jiazheng and Ji, Mingming and Zhao, Zhe and Bai, Ting},
  booktitle={Proceedings of the 2025 Conference on Empirical Methods in Natural Language Processing},
  pages={25972--25981},
  year={2025}
}

@article{r20,
  title={MemGPT: towards LLMs as operating systems.},
  author={Packer, Charles and Fang, Vivian and Patil, Shishir\_G and Lin, Kevin and Wooders, Sarah and Gonzalez, Joseph\_E},
  year={2023},
  publisher={ArXiv}
}

@article{r21,
  title={Zep: a temporal knowledge graph architecture for agent memory},
  author={Rasmussen, Preston and Paliychuk, Pavlo and Beauvais, Travis and Ryan, Jack and Chalef, Daniel},
  journal={arXiv preprint arXiv:2501.13956},
  year={2025}
}

@inproceedings{r22,
  title={Memorybank: Enhancing large language models with long-term memory},
  author={Zhong, Wanjun and Guo, Lianghong and Gao, Qiqi and Ye, He and Wang, Yanlin},
  booktitle={Proceedings of the AAAI conference on artificial intelligence},
  volume={38},
  number={17},
  pages={19724--19731},
  year={2024}
}

@article{r23,
  title={Mirix: Multi-agent memory system for llm-based agents},
  author={Wang, Yu and Chen, Xi},
  journal={arXiv preprint arXiv:2507.07957},
  year={2025}
}

@inproceedings{r24,
  title={Memoro: Using large language models to realize a concise interface for real-time memory augmentation},
  author={Zulfikar, Wazeer Deen and Chan, Samantha and Maes, Pattie},
  booktitle={Proceedings of the 2024 CHI Conference on Human Factors in Computing Systems},
  pages={1--18},
  year={2024}
}

@article{r25,
  title={From Verbatim to Gist: Distilling Pyramidal Multimodal Memory via Semantic Information Bottleneck for Long-Horizon Video Agents},
  author={Lian, Niu and Wang, Yuting and Yao, Hanshu and Wang, Jinpeng and Chen, Bin and Wang, Yaowei and Zhang, Min and Xia, Shu-Tao},
  journal={arXiv preprint arXiv:2603.01455},
  year={2026}
}

@article{r26,
  title={Token merging: Your vit but faster},
  author={Bolya, Daniel and Fu, Cheng-Yang and Dai, Xiaoliang and Zhang, Peizhao and Feichtenhofer, Christoph and Hoffman, Judy},
  journal={arXiv preprint arXiv:2210.09461},
  year={2022}
}

@inproceedings{r27,
  title={D$^2$Pruner: Debiased Importance and Structural Diversity for MLLM Token Pruning},
  author={Zhang, Evelyn and Yu, Fufu and Wu, Aoqi and Wen, Zichen and Yan, Ke and Ding, Shouhong and Qi, Biqing and Zhang, Linfeng},
  booktitle={Proceedings of the AAAI Conference on Artificial Intelligence},
  volume={40},
  number={15},
  pages={12412--12420},
  year={2026}
}

@inproceedings{r28,
  title={Llava-prumerge: Adaptive token reduction for efficient large multimodal models},
  author={Shang, Yuzhang and Cai, Mu and Xu, Bingxin and Lee, Yong Jae and Yan, Yan},
  booktitle={Proceedings of the IEEE/CVF International Conference on Computer Vision},
  pages={22857--22867},
  year={2025}
}

@inproceedings{r29,
  title={Hired: Attention-guided token dropping for efficient inference of high-resolution vision-language models},
  author={Arif, Kazi Hasan Ibn and Yoon, JinYi and Nikolopoulos, Dimitrios S and Vandierendonck, Hans and John, Deepu and Ji, Bo},
  booktitle={Proceedings of the AAAI Conference on Artificial Intelligence},
  volume={39},
  number={2},
  pages={1773--1781},
  year={2025}
}

@inproceedings{r30,
  title={Some methods of classification and analysis of multivariate observations},
  author={McQueen, James B},
  booktitle={Proc. of 5th berkeley symposium on math. stat. and prob.},
  pages={281--297},
  year={1967}
}

@article{r31,
  title={On the shortest spanning subtree of a graph and the traveling salesman problem},
  author={Kruskal, Joseph B},
  journal={Proceedings of the American Mathematical society},
  volume={7},
  number={1},
  pages={48--50},
  year={1956},
  publisher={JSTOR}
}

@article{r32,
  title={Streamforest: Efficient online video understanding with persistent event memory},
  author={Zeng, Xiangyu and Qiu, Kefan and Zhang, Qingyu and Li, Xinhao and Wang, Jing and Li, Jiaxin and Yan, Ziang and Tian, Kun and Tian, Meng and Zhao, Xinhai and others},
  journal={arXiv preprint arXiv:2509.24871},
  year={2025}
}

@inproceedings{r33,
  title={Timechat-online: 80\% visual tokens are naturally redundant in streaming videos},
  author={Yao, Linli and Li, Yicheng and Wei, Yuancheng and Li, Lei and Ren, Shuhuai and Liu, Yuanxin and Ouyang, Kun and Wang, Lean and Li, Shicheng and Li, Sida and others},
  booktitle={Proceedings of the 33rd ACM International Conference on Multimedia},
  pages={10807--10816},
  year={2025}
}

@article{r34,
  title={StreamingTOM: Streaming Token Compression for Efficient Video Understanding},
  author={Chen, Xueyi and Tao, Keda and Shao, Kele and Wang, Huan},
  journal={arXiv preprint arXiv:2510.18269},
  year={2025}
}

@article{r35,
  title={Recurrent Attention-based Token Selection for Efficient Streaming Video-LLMs},
  author={Dorovatas, Vaggelis and Seifi, Soroush and Gupta, Gunshi and Aljundi, Rahaf},
  journal={arXiv preprint arXiv:2510.17364},
  year={2025}
}

@inproceedings{r36,
  title={Online Video Understanding: OVBench and VideoChat-Online},
  author={Huang, Zhenpeng and Li, Xinhao and Li, Jiaqi and Wang, Jing and Zeng, Xiangyu and Liang, Cheng and Wu, Tao and Chen, Xi and Li, Liang and Wang, Limin},
  booktitle={Proceedings of the Computer Vision and Pattern Recognition Conference},
  pages={3328--3338},
  year={2025}
}

@inproceedings{r37,
  title={Videollm-online: Online video large language model for streaming video},
  author={Chen, Joya and Lv, Zhaoyang and Wu, Shiwei and Lin, Kevin Qinghong and Song, Chenan and Gao, Difei and Liu, Jia-Wei and Gao, Ziteng and Mao, Dongxing and Shou, Mike Zheng},
  booktitle={Proceedings of the IEEE/CVF Conference on Computer Vision and Pattern Recognition},
  pages={18407--18418},
  year={2024}
}

@article{r38,
  title={StreamMeCo: Long-Term Agent Memory Compression for Efficient Streaming Video Understanding},
  author={Wang, Junxi and Sun, Te and Zhu, Jiayi and Li, Junxian and Xu, Haowen and Wen, Zichen and Hu, Xuming and Li, Zhiyu and Zhang, Linfeng},
  journal={arXiv preprint arXiv:2604.09000},
  year={2026}
}

@inproceedings{r39,
  title={Evaluating very long-term conversational memory of llm agents},
  author={Maharana, Adyasha and Lee, Dong-Ho and Tulyakov, Sergey and Bansal, Mohit and Barbieri, Francesco and Fang, Yuwei},
  booktitle={Proceedings of the 62nd Annual Meeting of the Association for Computational Linguistics (Volume 1: Long Papers)},
  pages={13851--13870},
  year={2024}
}

@article{r40,
  title={Longmemeval: Benchmarking chat assistants on long-term interactive memory},
  author={Wu, Di and Wang, Hongwei and Yu, Wenhao and Zhang, Yuwei and Chang, Kai-Wei and Yu, Dong},
  journal={arXiv preprint arXiv:2410.10813},
  year={2024}
}

@article{r41,
  title={Know me, respond to me: Benchmarking llms for dynamic user profiling and personalized responses at scale},
  author={Jiang, Bowen and Hao, Zhuoqun and Cho, Young-Min and Li, Bryan and Yuan, Yuan and Chen, Sihao and Ungar, Lyle and Taylor, Camillo J and Roth, Dan},
  journal={arXiv preprint arXiv:2504.14225},
  year={2025}
}

@misc{r42,
      title={Qwen2.5 Technical Report}, 
      author={Qwen and : and An Yang and Baosong Yang and Beichen Zhang and Binyuan Hui and Bo Zheng and Bowen Yu and Chengyuan Li and Dayiheng Liu and Fei Huang and Haoran Wei and Huan Lin and Jian Yang and Jianhong Tu and Jianwei Zhang and Jianxin Yang and Jiaxi Yang and Jingren Zhou and Junyang Lin and Kai Dang and Keming Lu and Keqin Bao and Kexin Yang and Le Yu and Mei Li and Mingfeng Xue and Pei Zhang and Qin Zhu and Rui Men and Runji Lin and Tianhao Li and Tianyi Tang and Tingyu Xia and Xingzhang Ren and Xuancheng Ren and Yang Fan and Yang Su and Yichang Zhang and Yu Wan and Yuqiong Liu and Zeyu Cui and Zhenru Zhang and Zihan Qiu},
      year={2025},
      eprint={2412.15115},
      archivePrefix={arXiv},
      primaryClass={cs.CL},
      url={https://arxiv.org/abs/2412.15115}, 
}

@misc{r43,
      title={Beyond RAG for Agent Memory: Retrieval by Decoupling and Aggregation}, 
      author={Zhanghao Hu and Qinglin Zhu and Di Liang and Hanqi Yan and Yulan He and Lin Gui},
      year={2026},
      eprint={2602.02007},
      archivePrefix={arXiv},
      primaryClass={cs.CL},
      url={https://arxiv.org/abs/2602.02007}, 
}

@article{r45,
  title={Hierarchical event segmentation of episodic memory in virtual reality},
  author={Li, Yue and Johansson, Mikael and Nikolaev, Andrey R},
  journal={npj Science of Learning},
  volume={10},
  number={1},
  pages={25},
  year={2025},
  publisher={Nature Publishing Group UK London}
}

@article{r46,
  title={Event Segmentation Interventions Improve Memory for Naturalistic Events},
  author={Smith, Maverick E and Zacks, Jeffrey M},
  journal={Current Directions in Psychological Science},
  volume={35},
  number={1},
  pages={33--40},
  year={2026},
  publisher={SAGE Publications Sage CA: Los Angeles, CA}
}

@inproceedings{r47,
  title={Recipes for building an open-domain chatbot},
  author={Roller, Stephen and Dinan, Emily and Goyal, Naman and Ju, Da and Williamson, Mary and Liu, Yinhan and Xu, Jing and Ott, Myle and Smith, Eric Michael and Boureau, Y-Lan and others},
  booktitle={Proceedings of the 16th Conference of the European Chapter of the Association for Computational Linguistics: Main Volume},
  pages={300--325},
  year={2021}
}

@article{r48,
  title={Language models are few-shot learners},
  author={Brown, Tom and Mann, Benjamin and Ryder, Nick and Subbiah, Melanie and Kaplan, Jared D and Dhariwal, Prafulla and Neelakantan, Arvind and Shyam, Pranav and Sastry, Girish and Askell, Amanda and others},
  journal={Advances in neural information processing systems},
  volume={33},
  pages={1877--1901},
  year={2020}
}

@inproceedings{r49,
  title={Planning-driven programming: A large language model programming workflow},
  author={Lei, Chao and Chang, Yanchuan and Lipovetzky, Nir and Ehinger, Krista A},
  booktitle={Proceedings of the 63rd Annual Meeting of the Association for Computational Linguistics (Volume 1: Long Papers)},
  pages={12647--12684},
  year={2025}
}

@inproceedings{r50,
  title={Llm reasoning engine: Specialized training for enhanced mathematical reasoning},
  author={Chen, Shuguang and Lin, Guang},
  booktitle={Proceedings of the 4th International Workshop on Knowledge-Augmented Methods for Natural Language Processing},
  pages={118--128},
  year={2025}
}

@article{r51,
  title={TRUSTMEM: Learning Trustworthy Memory Consolidation for LLM Agents with Long-Term Memory},
  author={Yang, Tianyu and Paul, Sudipta and Srinivasan, Vijay and Kulkarni, Vivek and Chappidi, Srinivas},
  journal={arXiv preprint arXiv:2606.25161},
  year={2026}
}

@article{r52,
  title={Doc-to-Atom: Learning to Compile and Compose Memory Atoms},
  author={Diao, Xingjian and Li, Wenbo and Saidutta, Yashas Malur and Amballa, Avinash and Valkov, Lazar and Chappidi, Srinivas},
  journal={arXiv preprint arXiv:2606.12400},
  year={2026}
}

@inproceedings{r53,
  title={Tailoring memory granularity for multi-hop reasoning over long contexts},
  author={Qing, Peijun and Diao, Xingjian and Ma, Chiyu and Hassanpour, Saeed and Vosoughi, Soroush},
  booktitle={Findings of the Association for Computational Linguistics: EACL 2026},
  pages={3648--3666},
  year={2026}
}

@inproceedings{r54,
  title={Difficulty-aware agentic orchestration for query-specific multi-agent workflows},
  author={Su, Jinwei and Lan, Qizhen and Xia, Yinghui and Sun, Lifan and Tian, Weiyou and Shi, Tianyu and He, Lewei},
  booktitle={Proceedings of the ACM Web Conference 2026},
  pages={2060--2070},
  year={2026}
}

@article{r55,
  title={UIPress: Bringing Optical Token Compression to UI-to-Code Generation},
  author={Dai, Dasen and Li, Shuoqi and Chen, Ronghao and Wang, Huacan and Wu, Biao and Lan, Qizhen},
  journal={arXiv preprint arXiv:2604.09442},
  year={2026}
}

@inproceedings{r56,
  title={Knowme-bench: Benchmarking person understanding for lifelong digital companions},
  author={Wu, Tingyu and Chen, Zhisheng and Weng, Ziyan and Wang, Shuhe and Zhang, Shuo and Hu, Sen and Wu, Silin and Lan, Qizhen and Wang, Huacan and Chen, Ronghao},
  booktitle={Proceedings of the 64th Annual Meeting of the Association for Computational Linguistics (Volume 1: Long Papers)},
  pages={30214--30238},
  year={2026}
}

@article{r57,
  title={The Path to Self-Evolving Clinical Systems: Scaling Medical Agents from Assistance to Autonomy},
  author={Zhu, Chunzheng and Tian, Lei and Tan, Bohan and Zhou, Ziqi and Sun, Yuxuan and Wang, Yijun and Lv, Chengchao and Wen, Yilin and He, Yijun and Lin, Jinghao and others},
  journal={arXiv preprint arXiv:2607.11175},
  year={2026}
}

@inproceedings{r58,
  title={Medeyes: Learning dynamic visual focus for medical progressive diagnosis},
  author={Zhu, Chunzheng and Lin, Yangfang and Chen, Shen and Wang, Yijun and Lin, Jianxin},
  booktitle={Proceedings of the AAAI Conference on Artificial Intelligence},
  volume={40},
  number={16},
  pages={13916--13924},
  year={2026}
}

@article{r59,
  author       = {Xiaoyuan Li and
                  Yubo Ma and
                  Chengpeng Li and
                  Fengbin Zhu and
                  Yiyao Yu and
                  Keqin Bao and
                  Wenjie Wang and
                  Fuli Feng and
                  Dayiheng Liu},
  title        = {Unified Data Selection for {LLM} Reasoning},
  journal      = {CoRR},
  volume       = {abs/2605.22389},
  year         = {2026},
  url          = {https://doi.org/10.48550/arXiv.2605.22389},
  doi          = {10.48550/ARXIV.2605.22389},
  eprinttype   = {arXiv},
  eprint       = {2605.22389},
  bibsource    = {dblp computer science bibliography, https://dblp.org}
}

@article{r60,
  title={SAGE: Scalable Automated Robustness Augmentation for LLM Knowledge Evaluation},
  author={Li, Xiaoyuan and Wang, Yuzhe and Li, Moxin and Bao, Keqin and Men, Rui and Zhang, Yichang and Liu, Dayiheng and Wang, Wenjie and Feng, Fuli},
  journal={arXiv preprint arXiv:2605.12022},
  year={2026}
}

@article{r61,
  title={Vision-deepresearch benchmark: Rethinking visual and textual search for multimodal large language models},
  author={Zeng, Yu and Huang, Wenxuan and Fang, Zhen and Chen, Shuang and Shen, Yufan and Cai, Yishuo and Wang, Xiaoman and Yin, Zhenfei and Chen, Lin and Chen, Zehui and others},
  journal={arXiv preprint arXiv:2602.02185},
  year={2026}
}

@misc{r62,
      title={Video-DeepResearch: Towards the Next-Generation Multimodal Deepresearch Agent}, 
      author={Zhen Fang and Yu Zeng and Wenxuan Huang and Yiming Zhao and Shiting Huang and Tianfei Ren and Qi Lu and Qingnan Ren and Qisheng Su and Lionel Z. Wang and Qingyu Yin and Shuang Chen and Zehui Chen and Lin Chen and Zhenfei Yin and Yao Hu and Shaohui Lin and Wanli Ouyang and Shaosheng Cao and Feng Zhao},
      year={2026},
      eprint={2608.03979},
      archivePrefix={arXiv},
      primaryClass={cs.CV},
      url={https://arxiv.org/abs/2608.03979}, 
}


\clearpage
\appendix
\addtocontents{toc}{\protect\appendixTOCstart}
\renewcommand{\contentsname}{Appendix}

\begingroup
\setcounter{tocdepth}{2}
\makeatletter

\let\oldl@subsection\l@subsection
\renewcommand{\l@subsection}[2]{%
  \vspace{0.35ex}
  \oldl@subsection{#1}{#2}%
}

\makeatother
\tableofcontents
\endgroup

\section{Other Experimental Details}

\subsection{Baselines}
\label{appendixa.1}

\paragraph{Mem0.} Mem0 \cite{r1} is a long-term memory framework for large language models that addresses the limitation of fixed context windows by dynamically extracting, consolidating, and retrieving key information from conversations. It consists of two stages, memory extraction and memory update, and can automatically decide whether to add, update or delete information to maintain consistency and effectiveness of memory. Overall, it significantly reduces computational overhead while preserving reasoning ability, enabling AI agents to achieve stronger long-term interaction capability.

\paragraph{M3-Agent.} M3-Agent \cite{r7} is a multimodal agent framework that continuously perceives video and audio inputs and builds long-term memory to accumulate environmental knowledge. Its memory is organized in an entity-centric manner, integrating visual, auditory, and textual information to achieve more consistent understanding and representation. During task execution, it performs multi-turn reasoning and memory retrieval collaboratively to handle complex tasks more effectively. Overall, it enables multimodal agents to achieve more human-like perception, memory, and reasoning capabilities.

\paragraph{Random Pruning.} Regarding Random Pruning, we randomly select memory nodes for pruning according to the specified compression ratio.

\paragraph{KMeans.} Regarding KMeans \cite{r30}, we cluster the memory nodes using the KMeans algorithm and according to the desired compression ratio, partition them into a set of clusters. From each cluster, the memory node closest to the centroid is retained, while the remaining nodes are pruned.

\paragraph{DART.} DART \cite{r12} is a diversity-based token pruning approach that reduces computational overhead by identifying and removing highly redundant visual tokens. Its core idea is to select a small set of pivot tokens and preferentially retain tokens with low similarity to them, thereby preserving key information during compression. The method requires no additional training and is compatible with efficient attention mechanisms, achieving strong performance even under aggressive token reduction. We extend it to the agent memory compression setting: following the original setup, we randomly select 2\% of tokens as pivot tokens, compute the overall similarity between remaining tokens and these pivots, and retain those with the lowest similarity.

\paragraph{StreamMeCo.} StreamMeCo \cite{r38} is a memory compression method specifically designed for M3-Agent \cite{r7}, addressing the storage and retrieval bottlenecks caused by the growth of memory graphs in streaming video scenarios by efficiently compressing memory nodes in a structure-aware manner. Its core adopts a dual-branch strategy, performing representative sampling for isolated nodes and pruning connected nodes by jointly considering entity importance and semantic similarity, thereby preserving key information during compression.

\paragraph{Random Merging.} Regarding Random Merging, we partition the memory nodes into two sets according to the compression ratio, where the number of nodes in the second set corresponds to the number of nodes to be retained. We then randomly merge nodes from the second set with those in the first set until the target compression ratio is achieved.

\paragraph{ToMe.} ToMe \cite{r26} is a similarity-based token merging approach that reduces computational cost by progressively merging similar tokens within Transformer layers. Its core idea is to compute pairwise similarity between tokens and perform matching, merging similar tokens into a single representation, thereby preserving key information while reducing the number of tokens. The method can be applied directly without additional training and achieves a favorable trade-off between efficiency and performance across multimodal tasks. We extend it to the agent memory compression setting: following the original design, memory nodes are partitioned into two equally sized sets in an alternating manner based on their insertion order, and for each node in the first set, the most similar node in the second set is identified and merged. Each partitioning step can compress up to 50\% of the historical memory; if the target compression ratio is not reached, the process is iteratively repeated.

\subsection{Datasets}
\label{appendixa.2}

\paragraph{LoCoMo.} The LoCoMo dataset \cite{r39} consists of 10 very long conversations, each with about 27 sessions, 600 turns, and 16K tokens, designed to simulate long-term interactions. It includes 1,986 question answering instances across five categories, including single hop, multi hop, temporal reasoning, commonsense reasoning, and adversarial questions, to evaluate models’ long term memory and reasoning abilities.

\paragraph{LongMemEval.} The LongMemEval dataset \cite{r40} is a benchmark for evaluating long-term interactive memory, consisting of 500 user–assistant multi-session dialogue histories corresponding to 500 questions. Each dialogue is composed of multiple sessions, and the questions are categorized into six types, including single-session-user, single-session-assistant, single-session-preference, multi-session, knowledge-update, and temporal-reasoning, to assess models’ long-term memory and reasoning capabilities.

\paragraph{PersonaMem.} The PersonaMem dataset \cite{r41} provides multiple context-scale versions, including 32k, 128k, and 1M tokens. In this work, we adopt the PersonaMem-32k version, which contains 222 user–model multi-session dialogue histories and 589 questions. The question types mainly cover user fact recall, new idea generation, latest preference identification, preference evolution tracking, reasoning about preference changes, preference-aligned recommendation, and cross-scenario generalization, to evaluate models’ long-term personalized memory and reasoning capabilities.

\paragraph{M3-Bench-robot.} The M3-Bench-robot dataset \cite{r7} is an online video dataset consisting of 100 first-person robot videos across seven daily environments such as living rooms, kitchens, and offices, with a total of 1,276 questions. Each video involves interactions between the robot and multiple humans, requiring long-term memory construction and reasoning. The dataset includes five task types: multi-evidence reasoning, multi-hop reasoning, cross-modal reasoning, person understanding, and general knowledge extraction, to evaluate memory and reasoning in dynamic interactions.

\paragraph{M3-Bench-web.} The M3-Bench-web dataset \cite{r7} is an offline video dataset with 920 YouTube videos across 46 categories such as documentaries, travel, and sports, containing 3,214 questions. It includes five task types: multi-evidence reasoning, multi-hop reasoning, cross-modal reasoning, person understanding, and general knowledge extraction, to evaluate models’ cross-modal understanding and long-term memory reasoning.

\subsection{Additional Implementation Details}
\label{appendixa.3}

During the experiments, all settings for both the Mem0 and M3-Agent frameworks strictly follow their original papers \cite{r1, r7}. For the Mem0 framework, we use GPT-4o-mini to evaluate the overall answer quality, while for the M3-Agent framework, we employ GPT-4o for evaluation. All memory node merging processes are conducted using GPT-4o-mini. In terms of embedding models, Mem0 uses text-embedding-3-small, whereas M3-Agent adopts text-embedding-3-large. 

In addition, in Section~\ref{section5.3}, we use a sliding window approach: all memory nodes are sorted chronologically with a window size of 5. If the average embedding similarity between two adjacent windows exceeds 0.5, they are assigned to the same EventTree; otherwise, a new EventTree is initiated starting from the latter window.

\section{Supplementary Theoretical Analysis}

\subsection{Theoretical Analysis of Why Similar Memory Nodes Should Be Merged}
\label{appendixb.1}

In this section, we provide a detailed derivation of Eqs.~\ref{eq13}-\ref{eq15} in Section~\ref{section4} for completeness. To facilitate the following analysis, we first introduce two basic lemmas.

\textbf{Lemma 1 (Triangle Inequality).} For any vectors \(x\) and \(y\), we have
\begin{equation}
\|x + y\| \le \|x\| + \|y\|.
\label{eq16}
\end{equation}
\textbf{Lemma 2 (Cauchy--Schwarz Inequality).} For any vectors \(x\) and \(y\), we have
\begin{equation}
x^\top y \le \|x\| \cdot \|y\|.
\label{eq17}
\end{equation}
\noindent \textbf{Proof:}

By definition, the similarity between the merged memory node \( m_l \) and the query \( q \) is given by:
\begin{equation}
s_l = \frac{\lambda s_i + (1-\lambda)s_j}{\|\lambda e_i + (1-\lambda)e_j\|}.
\label{eq18}
\end{equation}
To bound the denominator, we apply Lemma 1, which gives:
\begin{equation}
\|\lambda e_i + (1-\lambda)e_j\| \le \lambda \|e_i\| + (1-\lambda)\|e_j\| = 1.
\label{eq19}
\end{equation}
Combining the above results, we obtain the following lower bound on \( s_l \):
\begin{equation}
s_l \ge \lambda s_i + (1-\lambda)s_j,
\label{eq20}
\end{equation}
which establishes Eq.~\ref{eq13}.

To further analyze the relationship between \( s_i \) and \( s_j \), we first rewrite their difference as:
\begin{equation}
s_i - s_j = q^\top (e_i - e_j).
\label{eq21}
\end{equation}
We then apply Lemma 2 to upper bound this term:
\begin{equation}
s_i - s_j \le \|q\| \cdot \|e_i - e_j\| = \|e_i - e_j\|.
\label{eq22}
\end{equation}
Next, we explicitly compute the norm of the difference between the two embeddings:
\begin{equation}
\|e_i - e_j\|^2 = \|e_i\|^2 + \|e_j\|^2 - 2 e_i^\top e_j = 2 - 2\rho_{ij}.
\label{eq23}
\end{equation}
Substituting this result back yields a lower bound for \( s_j \):
\begin{equation}
s_j \ge s_i - \sqrt{2 - 2\rho_{ij}},
\label{eq24}
\end{equation}
which establishes Eq.~\ref{eq14}.

Finally, by substituting Eq.~\ref{eq14} into Eq.~\ref{eq13}, we derive:
\begin{equation}
s_l \ge \lambda s_i + (1-\lambda)\left(s_i - \sqrt{2 - 2\rho_{ij}}\right).
\label{eq25}
\end{equation}
After simplification, we obtain the desired result:
\begin{equation}
s_l \ge s_i - (1-\lambda)\sqrt{2 - 2\rho_{ij}},
\label{eq26}
\end{equation}
which completes the proof of Eq.~\ref{eq15}.

\subsection{Theoretical Analysis of MemForest}
\label{appendixb.2}

In this section, we provide a detailed theoretical proof for MemForest.

\noindent \textbf{Proof:}

We analyze the memory merging strategy of MemForest. Specifically, MemForest compresses the historical memory set \( H = \{ m_i \}\mid_{i=1}^{N} \) into a compact set \( H' = \{ m'_j \}\mid_{j=1}^{M} \), where \( M < N \).

For each EventTree \( C_k \), as shown in Eq.~\ref{eq3}, the partitioning process jointly considers global semantic similarity and local temporal continuity. Therefore, at the initialization stage, the similarity between any two nodes \( m_i \) and \( m_j \) satisfies \( \rho_{ij} \ge \rho_{\min}^{(0)} \), where \( \rho_{\min}^{(0)} \) denotes the minimum similarity threshold under the initial partitioning. This threshold is typically maintained at a relatively high level.

During progressive merging, the minimum similarity dynamically evolves as new nodes are introduced. Let \( \rho_{\min}^{(t)} \) denote the minimum similarity after the \( t \)-th merge, then \( \rho_{\min}^{(t)} \le \rho_{\min}^{(t-1)} \). As indicated in Eq.~\ref{eq15}, MemForest always prioritizes merging the node pair with the highest similarity at each step, thereby keeping \( \rho_{\min} \) as large as possible after each merge and effectively mitigating the accumulation of semantic deviation.

According to Eq.~\ref{eq15}, at the \( t \)-th merging step, the similarity loss incurred by a single merge is bounded by:
\begin{equation}
s_i - s_l = \Delta s^{(t)} \le (1-\lambda)\sqrt{2 - 2\rho_{\min}^{(t)}}.
\label{eq27}
\end{equation}
For the \( k \)-th EventTree \( C_k \), let \( n_k \) and \( n'_k \) denote the numbers of memory nodes before and after compression, respectively. The total number of merging operations is:
\begin{equation}
n_k - n'_k.
\label{eq28}
\end{equation}

Denoting the similarity loss at each step as \( \Delta s^{(t)} \), the total similarity loss over the entire compression process can be written as:
\begin{equation}
\Delta S_k = \sum_{t=1}^{n_k - n'_k} \Delta s^{(t)}.
\label{eq29}
\end{equation}
Substituting Eq.~\ref{eq27} into the above equation, we obtain:
\begin{equation}
\Delta S_k \le \sum_{t=1}^{n_k - n'_k} (1-\lambda)\sqrt{2 - 2\rho_{\min}^{(t)}}.
\label{eq30}
\end{equation}
Since \( \rho_{\min}^{(t)} \) is monotonically non-increasing, we obtain the worst-case bound:
\begin{equation}
\Delta S_k \le (n_k - n'_k)(1-\lambda)\sqrt{2 - 2\rho_{\min}^{(f)}},
\label{eq31}
\end{equation}
where \( \rho_{\min}^{(f)} \) denotes the minimum similarity after the final merging step. 

Summing over all EventTrees yields the overall loss bound:
\begin{equation}
\Delta S \le (N - M)(1-\lambda)\sqrt{2 - 2\rho_{\min}^{(f)}}.
\label{eq32}
\end{equation}
Thus, the overall query relevance of the compressed set \( H' \) satisfies:
\begin{equation}
\sum_{m'_j \in H'} s'_j \ge \sum_{m_i \in H} s_i - (N - M)(1-\lambda)\sqrt{2 - 2\rho_{\min}^{(f)}}.
\label{eq33}
\end{equation}
This shows that the compression loss depends on the compression scale \( (N - M) \) and the minimum similarity within EventTrees. Since MemForest prioritizes merging highly similar nodes, \( \rho_{\min}^{(t)} \) can be consistently maintained at a relatively high level, effectively controlling error accumulation. As a result, MemForest maintains high query relevance even under high compression ratios.

\section{Memory Node Merging Prompt}

The prompt used in our memory node merging process is as follows:

\begin{tcolorbox}[colback=gray!5!white, colframe=black!50, title=Prompt Used in Memory Node Merging, sharp corners=southwest, fonttitle=\bfseries]
\baselineskip=1.60em
\textit{You are an intelligent memory summarization assistant. Please extract the core information from the following two memory entries and their timestamps, and generate a concise and coherent summary paragraph. If the memories contain relative time expressions (e.g., “last June,” “two months ago”), convert them into specific dates based on the corresponding timestamps (for example, if a memory mentions “last June” and the timestamp is February 2023, the actual time should be June 2022). The summary length should be kept within the combined length of the two original memories as much as possible. Output only the final summarized result, without any explanations, analysis, reasoning steps, or additional content.}\\
\textit{Memory 1: \{\}, Timestamp 1: \{\}}\\
\textit{Memory 2: \{\}, Timestamp 2: \{\}}
\end{tcolorbox}

\section{Additional Experiments}
\label{appendixd}

\subsection{Ablation Study under Different Ratios}

\vspace{-5pt}
\begin{table}[h]
\centering
\begin{minipage}{0.48\linewidth}
\centering
\caption{\textbf{Impact of different EventTree partitioning strategies at 30\% compression rate.}}
\vspace{-3pt}
\fontsize{7}{10}\selectfont
\setlength{\tabcolsep}{1.2pt}
\setlength{\abovecaptionskip}{4pt}
\setlength{\belowcaptionskip}{-10pt}
\begin{tabular}{l|ccc|c|c}
\toprule[1.2pt]
\textbf{Method} & \textbf{LoCoMo} & \textbf{LongMemEval} & \textbf{M3-Bench-web} & \textbf{Avg.} & \textbf{\%} \\
\midrule
\rowcolor{gray!7}
Baseline & 60.2 & 55.2 & 47.9 & 54.4 & 100.0\% \\
w/o \(g_{c_i}^j\) & 56.7 & 51.4 & 46.0 & 51.4 & 94.5\% \\
w/o \(l_{c_i}^j\) & 58.1 & 53.0 & 47.0 & 52.7& 96.9\% \\
\rowcolor{green!5}
\textbf{Ours} & \textbf{58.5} & \textbf{54.2} & \textbf{47.2} & \textbf{53.3} & \textbf{98.0\%} \\
\bottomrule[1.2pt]
\end{tabular}
\label{table5}
\end{minipage}
\hspace{0.01\linewidth}
\begin{minipage}{0.48\linewidth}
\centering
\caption{\textbf{Impact of different memory node merging strategies at 30\% compression rate.}}
\vspace{-3pt}
\fontsize{7}{10}\selectfont
\setlength{\tabcolsep}{1.2pt}
\setlength{\abovecaptionskip}{4pt}
\setlength{\belowcaptionskip}{-10pt}
\begin{tabular}{l|ccc|c|c}
\toprule[1.2pt]
\textbf{Method} & \textbf{LoCoMo} & \textbf{LongMemEval} & \textbf{M3-Bench-web} & \textbf{Avg.} & \textbf{\%} \\
\midrule
\rowcolor{gray!7}
Baseline & 60.2 & 55.2 & 47.9 & 54.4 & 100.0\% \\
Random & 57.0 & 51.6 & 46.4 & 51.7 & 95.0\% \\
Minimum & 54.6 & 51.0 & 45.9 & 50.5 & 92.8\% \\
\rowcolor{green!5}
\textbf{Ours} & \textbf{58.5} & \textbf{54.2} & \textbf{47.2} & \textbf{53.3} & \textbf{98.0\%} \\
\bottomrule[1.2pt]
\end{tabular}
\label{table6}
\end{minipage}
\end{table}
\vspace{0pt}

\begin{table}[h]
\vspace{-12pt}
\centering
\begin{minipage}{0.48\linewidth}
\centering
\caption{\textbf{Impact of different EventTree partitioning strategies at 70\% compression rate.}}
\vspace{-3pt}
\fontsize{7}{10}\selectfont
\setlength{\tabcolsep}{1.2pt}
\setlength{\abovecaptionskip}{4pt}
\setlength{\belowcaptionskip}{-10pt}
\begin{tabular}{l|ccc|c|c}
\toprule[1.2pt]
\textbf{Method} & \textbf{LoCoMo} & \textbf{LongMemEval} & \textbf{M3-Bench-web} & \textbf{Avg.} & \textbf{\%} \\
\midrule
\rowcolor{gray!7}
Baseline & 60.2 & 55.2 & 47.9 & 54.4 & 100.0\% \\
w/o \(g_{c_i}^j\) & 51.4 & 49.6 & 45.1 & 48.7 & 89.5\% \\
w/o \(l_{c_i}^j\) & 53.6 & 51.8 & 46.0 & 50.5 & 92.8\% \\
\rowcolor{green!5}
\textbf{Ours} & \textbf{54.4} & \textbf{52.2} & \textbf{46.3} & \textbf{51.0} & \textbf{93.8\%} \\
\bottomrule[1.2pt]
\end{tabular}
\label{table7}
\end{minipage}
\hspace{0.01\linewidth}
\begin{minipage}{0.48\linewidth}
\centering
\caption{\textbf{Impact of different memory node merging strategies at 70\% compression rate.}}
\vspace{-3pt}
\fontsize{7}{10}\selectfont
\setlength{\tabcolsep}{1.2pt}
\setlength{\abovecaptionskip}{4pt}
\setlength{\belowcaptionskip}{-10pt}
\begin{tabular}{l|ccc|c|c}
\toprule[1.2pt]
\textbf{Method} & \textbf{LoCoMo} & \textbf{LongMemEval} & \textbf{M3-Bench-web} & \textbf{Avg.} & \textbf{\%} \\
\midrule
\rowcolor{gray!7}
Baseline & 60.2 & 55.2 & 47.9 & 54.4 & 100.0\% \\
Random & 51.5 & 50.8 & 45.3 & 49.2 & 90.4\% \\
Minimum & 49.8 & 48.0 & 44.2 & 47.3 & 87.0\% \\
\rowcolor{green!5}
\textbf{Ours} & \textbf{54.4} & \textbf{52.2} & \textbf{46.3} & \textbf{51.0} & \textbf{93.8\%} \\
\bottomrule[1.2pt]
\end{tabular}
\label{table8}
\end{minipage}
\end{table}
\vspace{-5pt}

As shown in Tables~\ref{table5}-\ref{table8}, when compressing \textbf{30\%} and \textbf{70\%} of historical memory, both the global similarity score \(g_{c_i}^j\) and the local continuity score \(l_{c_i}^j\) significantly improve performance. In addition, employing the other two merging strategies results in a notable performance drop, which is consistent with the previous experiments.

\subsection{Other Ablation Study}

\vspace{-3pt} 
\begin{table}[h]
\centering
\caption{\textbf{Effect of node degree at different compression ratios.}}
\vspace{-3pt}
\fontsize{7}{10}\selectfont
\setlength{\tabcolsep}{7pt}
\setlength{\abovecaptionskip}{4pt}
\setlength{\belowcaptionskip}{-10pt}
\begin{tabular}{l|ccc|c|c}
\toprule[1.2pt]
\textbf{Method} & \textbf{LoCoMo} & \textbf{LongMemEval} & \textbf{M3-Bench-web} & \textbf{Avg.} & \textbf{\%} \\
\midrule
\rowcolor{gray!7}
Baseline & 60.2 & 55.2 & 47.9 & 54.4 & 100\% \\

w/o \(\text{deg}(\cdot)\) ($\downarrow 30\%$) & 58.1 & 53.6 & 47.2 & 53.0 & 97.4\% \\
\rowcolor{green!4}
w/ \(\text{deg}(\cdot)\)  ($\downarrow 30\%$) & \textbf{58.5} & \textbf{54.2} & \textbf{47.2} & \textbf{53.3} & \textbf{98.0\%} \\

w/o \(\text{deg}(\cdot)\) ($\downarrow 50\%$) & 57.1 & 51.4 & \textbf{46.6} & 51.7 & 95.0\% \\
\rowcolor{green!7}
w/ \(\text{deg}(\cdot)\) ($\downarrow 50\%$)  & \textbf{58.4} & \textbf{51.8} & 46.5 & \textbf{52.2} & \textbf{96.0\%} \\

w/o \(\text{deg}(\cdot)\) ($\downarrow 70\%$) & 53.6 & 50.4 & 46.0 & 50.0 & 91.9\% \\
\rowcolor{green!10}
w/ \(\text{deg}(\cdot)\) ($\downarrow 70\%$)  & \textbf{54.4} & \textbf{52.2} & \textbf{46.3} & \textbf{51.0} & \textbf{93.8\%} \\
\bottomrule[1.2pt]
\end{tabular}
\label{table9}
\end{table}
\vspace{-5pt}

As shown in Table~\ref{table9}, we conducted ablation experiments on the effect of node degree \(\text{deg}(\cdot)\) under different compression ratios. Across three representative benchmarks, incorporating node degree consistently improves performance compared to not using it, indicating that in the maximum spanning tree, nodes with higher degrees are typically core nodes of the event, carry more critical information, and should not be merged prematurely.

\begin{table*}[htbp]
\vspace{-5pt}
\fontsize{8}{13}\selectfont
\renewcommand{\arraystretch}{0.8}
\setlength{\tabcolsep}{4.5pt}
\centering
\caption{\textbf{Performance comparison of different baselines on the LoCoMo benchmark in terms of B1 and F1 under varying compression ratios.} {\color{blue!30}Blue} denotes pruning methods, while {\color{red!30}Red} denotes merging methods. For each ratio, the top two results are highlighted in \textbf{bold black font}.}
\vspace{3pt}
\begin{tabular}{lcc|cc|cc|cc|cc}
\toprule[1.5pt]
\multicolumn{1}{l}{\textbf{Category}}
& \multicolumn{2}{c|}{SH} 
& \multicolumn{2}{c|}{MH} 
& \multicolumn{2}{c|}{TR} 
& \multicolumn{2}{c|}{OD} 
& \multicolumn{2}{c}{\textbf{All}} \\
\cmidrule(lr){2-3}\cmidrule(lr){4-5}\cmidrule(lr){6-7}\cmidrule(lr){8-9}\cmidrule(lr){10-11}
\multicolumn{1}{l}{\textbf{Method}}
& B1 & F1 
& B1 & F1 
& B1 & F1 
& B1 & F1 
& B1 & F1 \\
\midrule

  Mem0  &36.6&46.6&22.6&32.3&40.6&49.9&14.5&20.3&33.5&42.0 \\
  \hspace{7pt}+ AGPR
        &36.3&43.9&21.6&32.0&39.7&48.7&13.5&19.9&32.9&41.4 \\

  \rowcolor{gray!15}
  & \multicolumn{8}{c}{\emph{Historical memory compression ratio ({\color{green!40!black}$\downarrow 30\%$})}}
  & \multicolumn{2}{c}{~} \\

  \rowcolor{blue!3}
  \hspace{7pt}+ Random Pruning
        &29.5&36.3&20.6&30.7&34.6&40.9&14.5&17.5&28.0&35.1 \\
  \rowcolor{blue!3}
  \hspace{7pt}+ KMeans
        &33.4&40.9&21.4&32.0&39.5&46.7&14.6&20.1&31.3&39.3 \\
  \rowcolor{blue!3}
  \hspace{7pt}+ DART
        &34.4&41.9&22.1&32.1&35.9&43.4&13.8&19.4&31.2&39.0 \\
  \rowcolor{red!3}
  \hspace{7pt}+ Random Merging
        &32.7&39.9&20.0&29.7&35.6&46.4&16.3&21.7&30.0&38.3 \\
  \rowcolor{red!3}
  \hspace{7pt}+ ToMe
        &33.4&41.4&20.9&31.2&37.4&47.2&15.9&20.7&30.9&39.5 \\
  \rowcolor{red!3}
  \hspace{7pt}+ MemForest
        &33.8&41.6&22.6&33.8&37.1&46.7&18.0&23.4&\textbf{31.5}&\textbf{40.1} \\
  \rowcolor{red!3}
  \hspace{7pt}+ MemForest + AGPR
        &34.1&41.8&23.0&34.4&38.3&48.2&17.3&23.3&\textbf{31.9}&\textbf{40.6} \\

  \midrule

  \rowcolor{gray!15}
  & \multicolumn{8}{c}{\emph{Historical memory compression ratio ({\color{green!40!black}$\downarrow 50\%$})}}
  & \multicolumn{2}{c}{~} \\

  \rowcolor{blue!3}
  \hspace{7pt}+ Random Pruning
        &23.2&28.4&16.5&25.5&27.2&32.5&14.1&19.3&22.2&28.1 \\
  \rowcolor{blue!3}
  \hspace{7pt}+ KMeans
        &28.1&34.1&16.9&25.9&30.7&36.4&13.9&17.6&25.7&32.0 \\
  \rowcolor{blue!3}
  \hspace{7pt}+ DART
        &31.8&37.9&21.5&30.8&34.1&40.6&15.3&20.8&29.3&36.1 \\
  \rowcolor{red!3}
  \hspace{7pt}+ Random Merging
        &28.3&35.5&21.4&30.8&35.6&45.4&15.0&19.8&27.7&35.8 \\
  \rowcolor{red!3}
  \hspace{7pt}+ ToMe
        &31.1&38.3&20.8&30.4&35.0&45.4&15.2&19.7&29.1&37.2 \\
  \rowcolor{red!3}
  \hspace{7pt}+ MemForest
        &32.6&39.7&21.9&31.1&36.7&45.9&16.2&21.6&\textbf{30.5}&\textbf{38.3} \\
  \rowcolor{red!3}
  \hspace{7pt}+ MemForest + AGPR
        &32.6&40.0&23.9&33.7&36.4&45.7&15.4&20.8&\textbf{30.7}&\textbf{38.8} \\

  \midrule

  \rowcolor{gray!15}
  & \multicolumn{8}{c}{\emph{Historical memory compression ratio ({\color{green!40!black}$\downarrow 70\%$})}}
  & \multicolumn{2}{c}{~} \\
  
  \rowcolor{blue!3}
  \hspace{7pt}+ Random Pruning
        &18.0&22.6&13.9&21.2&21.4&25.1&11.2&15.1&17.5&22.4 \\
  \rowcolor{blue!3}
  \hspace{7pt}+ KMeans
        &20.7&25.7&14.4&23.3&23.7&27.5&14.9&20.7&19.9&25.3 \\
  \rowcolor{blue!3}
  \hspace{7pt}+ DART
        &25.6&31.0&18.9&27.6&28.4&34.1&13.6&19.1&24.2&30.3 \\
  \rowcolor{red!3}
  \hspace{7pt}+ Random Merging
        &27.0&33.5&19.6&29.6&36.8&45.4&17.4&22.3&27.1&34.6 \\
  \rowcolor{red!3}
  \hspace{7pt}+ ToMe
        &27.7&34.7&20.2&29.3&34.4&44.5&12.8&17.6&26.8&34.7 \\
  \rowcolor{red!3}
  \hspace{7pt}+ MemForest
        &29.3&35.8&23.6&32.9&34.9&42.5&14.2&20.1&\textbf{28.4}&\textbf{35.7} \\
  \rowcolor{red!3}
  \hspace{7pt}+ MemForest + AGPR
        &29.6&36.2&23.3&32.6&35.1&43.0&15.5&21.6&\textbf{28.7}&\textbf{36.0} \\

  \bottomrule[1.5pt]
\end{tabular}
\label{table10}
\end{table*}

\begin{figure*}[h]
  \centering  
  \includegraphics[width=\textwidth]{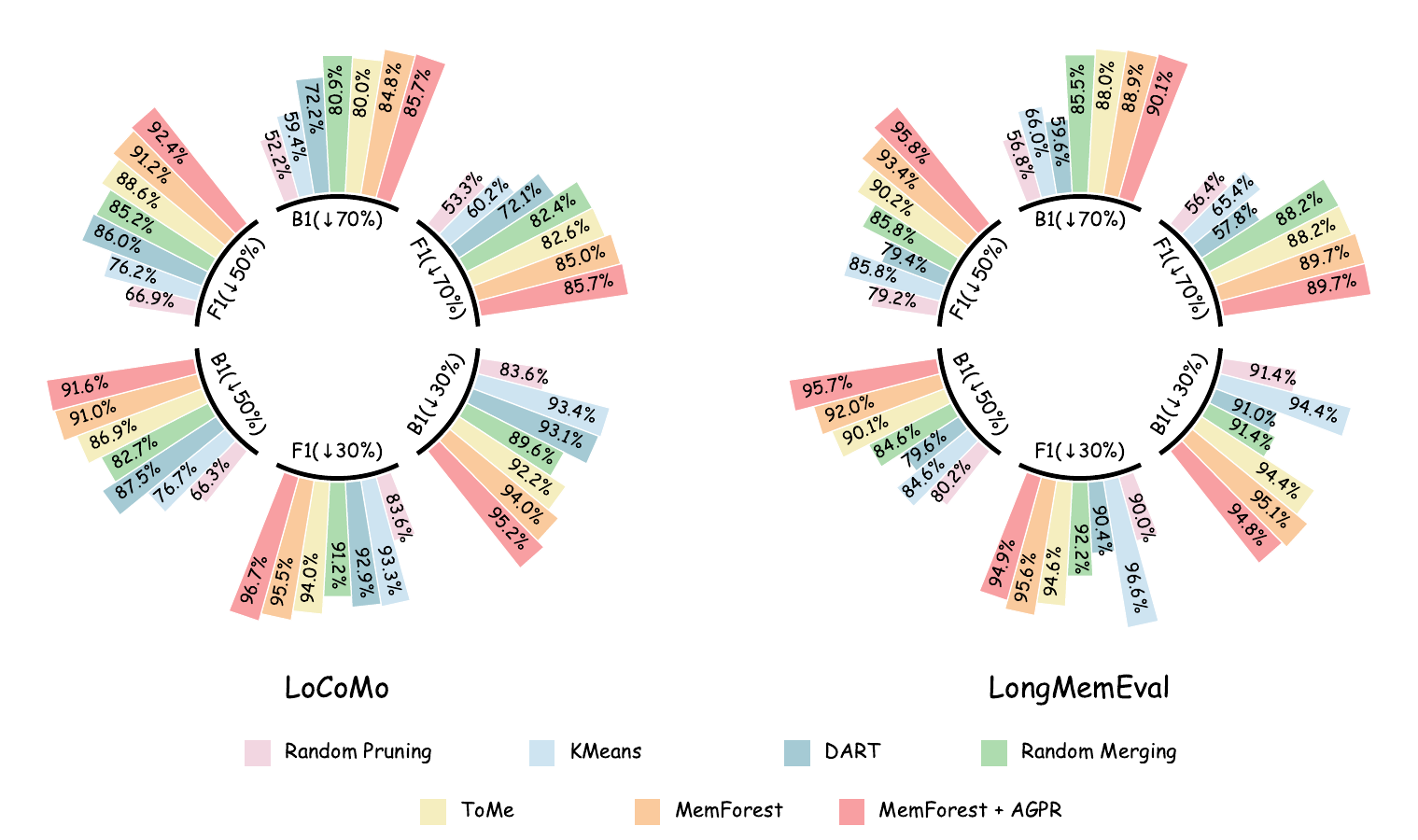}  
  \vspace{-20pt} 
  \caption{\textbf{Retention of B1 and F1 scores.} On two benchmarks, the percentages of B1 and F1 retained relative to the original performance across methods under varying compression rates.}
  \label{Fig4}
  \vspace{-12pt} 
\end{figure*}

\clearpage

\begin{table*}[h]
\fontsize{8}{13}\selectfont
\renewcommand{\arraystretch}{0.8}
\setlength{\tabcolsep}{2.8pt}
\centering
\caption{\textbf{Performance comparison of different baselines on the LongMemEval benchmark in terms of B1 and F1 under varying compression ratios.} {\color{blue!30}Blue} denotes pruning methods, while {\color{red!30}Red} denotes merging methods. For each ratio, the top two results are highlighted in \textbf{bold black font}.}
\vspace{3pt}
\begin{tabular}{lcc|cc|cc|cc|cc|cc|cc}
\toprule[1.5pt]
\multicolumn{1}{l}{\textbf{Category}}
& \multicolumn{2}{c|}{SU} 
& \multicolumn{2}{c|}{SA} 
& \multicolumn{2}{c|}{SP} 
& \multicolumn{2}{c|}{MS} 
& \multicolumn{2}{c|}{KU}
& \multicolumn{2}{c|}{TR} 
& \multicolumn{2}{c}{\textbf{All}} \\
\cmidrule(lr){2-3}\cmidrule(lr){4-5}\cmidrule(lr){6-7}\cmidrule(lr){8-9}\cmidrule(lr){10-11}\cmidrule(lr){12-13}\cmidrule(lr){14-15}
\multicolumn{1}{l}{\textbf{Method}}
& B1 & F1 
& B1 & F1 
& B1 & F1 
& B1 & F1 
& B1 & F1 
& B1 & F1 
& B1 & F1 \\
\midrule

  Mem0  &64.1&72.6&15.1&15.5&0.1&7.6&32.1&38.2&39.7&51.1&26.3&38.9&32.4&40.8 \\
  \hspace{7pt}+ AGPR
        &62.7&71.8&14.9&15.3&0.1&8.0&30.4&35.7&40.5&52.4&26.5&38.9&31.9&40.3 \\

  \rowcolor{gray!15}
  & \multicolumn{12}{c}{\emph{Historical memory compression ratio ({\color{green!40!black}$\downarrow 30\%$})}}
  & \multicolumn{2}{c}{~} \\

  \rowcolor{blue!3}
  \hspace{7pt}+ Random Pruning
        &58.7&67.2&14.3&13.3&0.1&7.8&27.2&32.0&35.3&44.7&26.3&37.2&29.6&36.7 \\
  \rowcolor{blue!3}
  \hspace{7pt}+ KMeans
        &60.1&59.4&13.8&14.4&0.1&7.7&29.4&35.7&39.6&51.4&24.8&38.0&30.6&\textbf{39.4} \\
  \rowcolor{blue!3}
  \hspace{7pt}+ DART
        &57.2&65.8&14.7&16.2&0.0&6.1&27.4&31.3&39.5&50.2&24.0&35.3&29.5&36.9 \\
  \rowcolor{red!3}
  \hspace{7pt}+ Random Merging
        &60.0&69.1&12.9&14.1&0.0&6.9&26.1&30.4&37.3&48.9&26.3&38.4&29.6&37.6 \\
  \rowcolor{red!3}
  \hspace{7pt}+ ToMe
        &59.3&68.6&14.8&15.8&0.0&7.1&29.7&34.2&37.0&47.9&26.3&38.3&30.6&38.6 \\
  \rowcolor{red!3}
  \hspace{7pt}+ MemForest
        &58.9&69.1&13.3&13.6&0.0&6.0&30.8&35.5&38.4&49.5&25.8&38.7&\textbf{30.8}&\textbf{39.0} \\
  \rowcolor{red!3}
  \hspace{7pt}+ MemForest + AGPR
        &60.4&69.0&13.2&13.3&0.0&6.4&30.1&35.2&38.7&50.2&25.2&37.5&\textbf{30.7}&38.7 \\

  \midrule

  \rowcolor{gray!15}
  & \multicolumn{12}{c}{\emph{Historical memory compression ratio ({\color{green!40!black}$\downarrow 50\%$})}}
  & \multicolumn{2}{c}{~} \\

  \rowcolor{blue!3}
  \hspace{7pt}+ Random Pruning
        &49.7&57.5&12.6&13.9&0.0&6.9&24.2&25.0&30.6&39.5&24.2&35.5&26.0&32.3 \\
  \rowcolor{blue!3}
  \hspace{7pt}+ KMeans
        &53.1&61.8&12.1&13.3&0.0&6.8&24.9&28.3&34.9&46.3&24.6&36.3&27.4&35.0 \\
  \rowcolor{blue!3}
  \hspace{7pt}+ DART
        &47.4&54.7&11.7&11.1&0.1&7.7&26.2&30.2&30.3&39.8&23.3&33.1&25.8&32.4 \\
  \rowcolor{red!3}
  \hspace{7pt}+ Random Merging
        &55.2&63.7&14.1&15.2&0.1&9.2&25.2&29.4&30.6&41.4&24.8&35.9&27.4&35.0 \\
  \rowcolor{red!3}
  \hspace{7pt}+ ToMe
        &56.5&66.5&13.2&13.4&0.0&7.2&29.6&34.1&31.8&41.8&26.3&37.3&29.2&36.8 \\
  \rowcolor{red!3}
  \hspace{7pt}+ MemForest
        &59.8&69.5&13.2&14.6&0.0&7.0&28.5&33.1&38.3&49.2&24.1&36.8&\textbf{29.8}&\textbf{38.1} \\
  \rowcolor{red!3}
  \hspace{7pt}+ MemForest + AGPR
        &60.6&70.2&15.2&16.4&0.0&7.6&28.7&33.3&38.8&49.0&26.6&39.4&\textbf{31.0}&\textbf{39.1} \\

  \midrule

  \rowcolor{gray!15}
  & \multicolumn{12}{c}{\emph{Historical memory compression ratio ({\color{green!40!black}$\downarrow 70\%$})}}
  & \multicolumn{2}{c}{~} \\
  
  \rowcolor{blue!3}
  \hspace{7pt}+ Random Pruning
        &30.3&36.1&7.0&7.6&0.1&8.0&15.6&14.0&20.8&27.0&22.3&32.6&18.4&23.0 \\
  \rowcolor{blue!3}
  \hspace{7pt}+ KMeans
        &44.0&51.0&9.6&8.0&0.0&6.6&15.0&15.3&29.0&37.4&21.3&31.5&21.4&26.7 \\
  \rowcolor{blue!3}
  \hspace{7pt}+ DART
        &28.4&32.7&10.1&9.3&0.1&5.4&20.7&21.9&22.3&30.5&19.4&26.7&19.3&23.6 \\
  \rowcolor{red!3}
  \hspace{7pt}+ Random Merging
        &54.1&65.2&17.2&19.3&0.1&8.3&25.3&30.6&32.3&43.8&24.0&34.7&27.7&36.0 \\
  \rowcolor{red!3}
  \hspace{7pt}+ ToMe
        &55.6&64.3&15.6&16.2&0.0&6.1&27.4&31.3&29.5&40.3&26.6&38.2&28.5&36.0 \\
  \rowcolor{red!3}
  \hspace{7pt}+ MemForest
        &54.2&63.9&16.1&17.1&0.0&7.3&28.5&32.8&34.7&44.0&24.2&36.4&\textbf{28.8}&\textbf{36.6} \\
  \rowcolor{red!3}
  \hspace{7pt}+ MemForest + AGPR
        &55.7&65.2&16.6&18.4&0.0&6.7&28.4&33.0&34.2&43.6&25.0&35.5&\textbf{29.2}&\textbf{36.6} \\

  \bottomrule[1.5pt]
\end{tabular}
\label{table11}
\vspace{-12pt} 
\end{table*}

\subsection{Other Evaluation Metrics}

Furthermore, for the Mem0 framework, we evaluated the \textbf{B1} and \textbf{F1} scores on two representative benchmarks: LoCoMo and LongMemEval. As shown in Tables~\ref{table10}–\ref{table11} and Figure~\ref{Fig4}, our method consistently outperforms other approaches across both benchmarks, further demonstrating the effectiveness of MemForest.

\subsection{Parameter Sensitivity Analysis}

\begin{figure*}[h]
  \centering  
  \includegraphics[width=\textwidth]{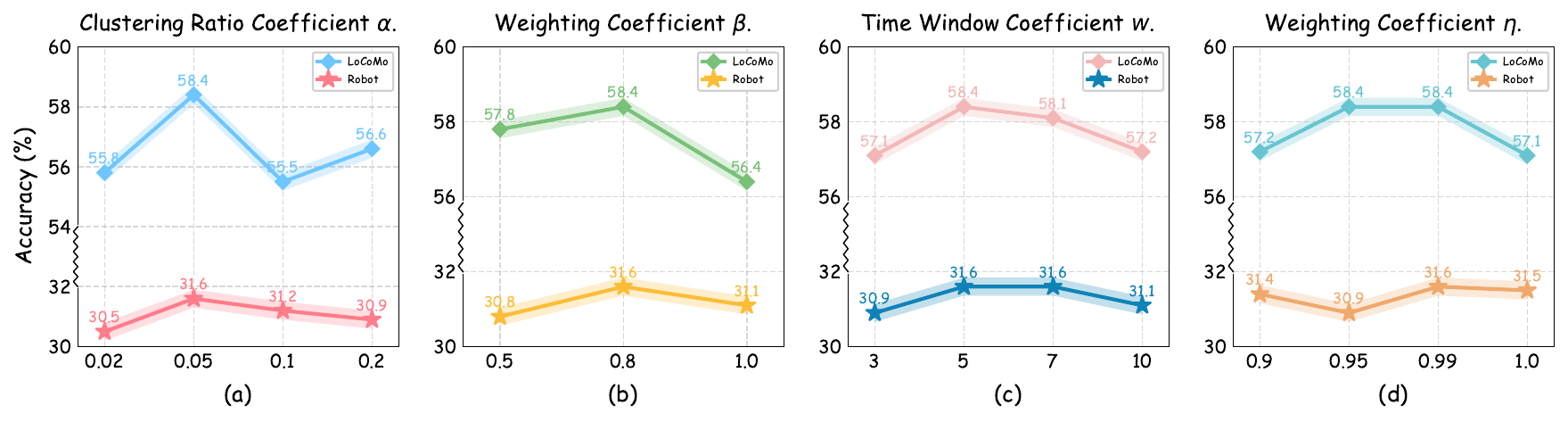}  
  \vspace{-20pt} 
  \caption{\textbf{Impact of Parameter Variations on Merging Performance.}}
  \label{Fig5}
  \vspace{-12pt} 
\end{figure*}
\vspace{10pt}

To analyze the sensitivity of different parameters, we conduct experiments on the LoCoMo and M3-Bench-robot datasets under a compression ratio of \textbf{50\%}.

\paragraph{Clustering Ratio Coefficient \(\alpha\).}

As shown in Fig.~\ref{Fig5}{\color{red}(a)}, when the clustering ratio coefficient \(\alpha\) is set to \textbf{0.05}, the performance on both benchmarks reaches its optimum. When \(\alpha\) is either too high or too low, the performance degrades. This indicates that when the number of EventTrees is too small, excessive unrelated memories are grouped within a single EventTree; whereas when the number is too large, memory nodes from the same event are overly fragmented, thereby harming overall performance.

\paragraph{Weighting Coefficient \(\beta\).}

As shown in Fig.~\ref{Fig5}\textcolor{red}{(b)}, when the weighting coefficient \(\beta\) is set to \textbf{0.8}, the performance on both benchmarks reaches its optimum. When \(\beta\) is too large, the effect of local temporal continuity is weakened; whereas when \(\beta\) is too small, temporal continuity becomes dominant, leading to a suppression of global semantic similarity, thereby degrading the overall performance.

\paragraph{Time Window Coefficient \(w\).}

As shown in Fig.~\ref{Fig5}\textcolor{red}{(c)}, when the time window coefficient \(w\) is set to \textbf{5}, the performance on both benchmarks reaches its optimum. When \(w\) is too large, each cluster contains too many neighboring memory nodes, which weakens the aggregation of semantically similar nodes; whereas when \(w\) is too small, the number of neighboring nodes is insufficient to provide adequate local temporal information, thereby affecting the overall performance.

\paragraph{Weighting Coefficient \(\eta\).}

As shown in Fig.~\ref{Fig5}\textcolor{red}{(d)}, the performance on both benchmarks reaches its optimum when the weighting coefficient \(\eta\) is set to \textbf{0.99}. When \(\eta\) is too small, the degree information dominates, which suppresses the merging of semantically similar nodes; whereas when \(\eta\) is set to \textbf{1}, the importance of node degrees is ignored, potentially causing core nodes to be merged prematurely.

\begin{figure*}[h]
  \centering  
  \includegraphics[width=\textwidth]{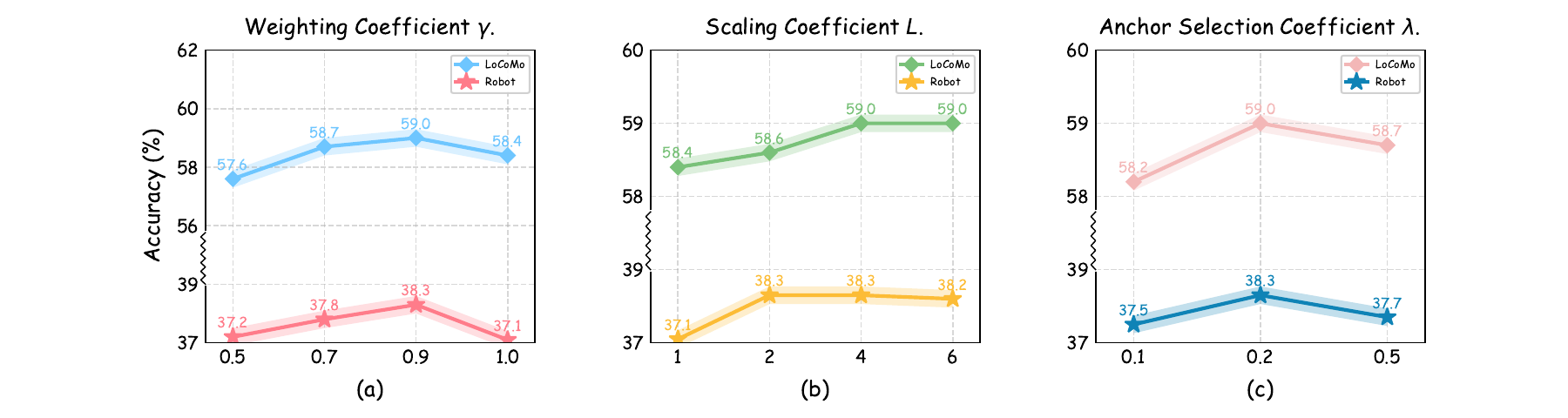}  
  \vspace{-20pt} 
  \caption{\textbf{Impact of Parameter Variations on Memory Retrieval.}}
  \label{Fig6}
  \vspace{-12pt} 
\end{figure*}
\vspace{5pt}

\paragraph{Weighting Coefficient \(\gamma\).}

As shown in Fig.~\ref{Fig6}\textcolor{red}{(a)}, when the weighting coefficient \(\gamma\) is set to \(\textbf{0.9}\), the performance on both benchmarks reaches its optimum. When \(\gamma\) is too low, the effect of semantic similarity on retrieval is weakened, leading to degraded performance; whereas when \(\gamma = \textbf{1}\), retrieval relies solely on semantic similarity, ignoring the neighborhood information of key memory nodes, which results in performance degradation.

\paragraph{Scaling Coefficient \(L\).}

As shown in Fig.~\ref{Fig6}\textcolor{red}{(b)}, when the scaling coefficient \(L\) is set to \textbf{4}, the performance on both benchmarks reaches its optimum. When \(L\) is too small, the number of recalled candidate memory nodes is insufficient to provide adequate temporal neighborhood compensation, thereby affecting retrieval performance; whereas when \(L\) is greater than \textbf{4}, the performance remains almost unchanged, as the number of candidate memory nodes is sufficient to fully cover the neighborhood of key memory nodes.

\paragraph{Anchor Selection Coefficient \(\lambda\).}

As shown in Fig.~\ref{Fig6}\textcolor{red}{(c)}, when the anchor selection coefficient \(\lambda\) is set to \textbf{0.2}, the performance on both benchmarks reaches its optimum. When \(\lambda\) is too low, the number of anchors is insufficient, resulting in inadequate propagation of neighborhood information and preventing some relevant memory nodes from being effectively retrieved; whereas when \(\lambda\) is too high, excessive anchors introduce redundant information and additional noise, which negatively affects retrieval performance.

\subsection{Memory Node Merging Cost Analysis}

Regarding the cost of memory node merging, each merge only requires a short prompt and two memory entries as input, producing a single summarized memory, resulting in very low overall cost. When using GPT-4o-mini to merge memory nodes and compress \textbf{50\%} of the historical memory, the cost is only about \textbf{\$0.1} for the LoCoMo benchmark ($\approx$100MB, 6{,}000 memory nodes) or roughly two hours of streaming video, making the practical overhead negligible. Moreover, this process can also be implemented with locally deployed open-source models.

\subsection{Merging Performance across Different Models}
\label{appendixd.6}

We evaluate Qwen2.5-7B-Instruct \cite{r42}, GPT-4o-mini, and GPT-5.2 under a \textbf{50\%} compression ratio on the LoCoMo benchmark. As shown in Table~\ref{table12}, GPT-4o-mini and GPT-5.2 achieve comparable merging performance when used as backbone models, suggesting that once the model size reaches a certain scale, the performance tends to saturate with limited room for further improvement. In contrast, Qwen2.5-7B-Instruct performs worse, likely due to its smaller parameter size. This observation also raises an interesting question: can a dedicated lightweight model be trained specifically for memory merging, and could it achieve performance comparable to or even surpass that of models like GPT-4o-mini?

\vspace{-3pt} 
\begin{table}[h]
\centering
\caption{\textbf{Merging performance of different models.}}
\vspace{-3pt}
\fontsize{7}{10}\selectfont
\setlength{\tabcolsep}{8pt}
\setlength{\abovecaptionskip}{4pt}
\setlength{\belowcaptionskip}{-10pt}
\begin{tabular}{l|c|c}
\toprule[1.2pt]
\textbf{Method} & \textbf{LoCoMo} & \textbf{\%} \\
\midrule
\rowcolor{gray!7}
Baseline & 60.2 & 100.0\% \\
\rowcolor{green!1}
Qwen2.5-7B-Instruct & 52.9 & 87.9\% \\
\rowcolor{green!5}
GPT-4o-mini & \textbf{58.4} & \textbf{97.0\%} \\
\rowcolor{green!9}
GPT-5.2 & 58.1 & 96.5\% \\
\bottomrule[1.2pt]
\end{tabular}
\label{table12}
\end{table}
\vspace{-5pt} 

\subsection{Detailed time efficiency results}
\label{appendixd.7}

\vspace{-3pt} 
\begin{table}[h]
\centering
\caption{\textbf{\textit{T-Time(s)} across the LoCoMo, LongMemEval, and PersonaMem benchmarks.}}
\vspace{-3pt}
\fontsize{7}{10}\selectfont
\setlength{\tabcolsep}{7pt}
\setlength{\abovecaptionskip}{4pt}
\setlength{\belowcaptionskip}{-10pt}
\begin{tabular}{l|c|c|c}
\toprule[1.2pt]
\textbf{Dataset} & \textbf{LoCoMo} & \textbf{LongMemEval} & \textbf{PersonaMem} \\
\midrule
\rowcolor{gray!7}
Baseline & 0.48 & 8.30 & 4.90 \\
\rowcolor{green!1}
Baseline + AGPR & 0.48 & 8.31 & 4.90 \\
\rowcolor{green!4}
Baseline + MemForest + AGPR ( $\downarrow 30\%$ ) 
& 0.30 & 6.04 & 3.56 \\
\rowcolor{green!7}
Baseline + MemForest + AGPR ( $\downarrow 50\%$ ) & 0.24 & 4.32 & 2.78 \\
\rowcolor{green!10}
Baseline + MemForest + AGPR ( $\downarrow 70\%$ ) & 0.16 & 2.70 & 1.34 \\
\bottomrule[1.2pt]
\end{tabular}
\label{table13}
\end{table}
\vspace{-5pt}

\vspace{-3pt} 
\begin{table}[h]
\centering
\caption{\textbf{\textit{Avg-R(×), R-Time(s), and T-Time(s)} across M3-Bench-robot and M3-Bench-web.}}
\vspace{-5pt}
\fontsize{7}{10}\selectfont
\setlength{\tabcolsep}{5pt}
\setlength{\abovecaptionskip}{4pt}
\setlength{\belowcaptionskip}{-10pt}
\begin{tabular}{l|ccc|ccc}
\toprule[1.2pt]
\multirow{2}{*}{\textbf{Dataset}} 
& \multicolumn{3}{c|}{\textbf{M3-Bench-robot}} 
& \multicolumn{3}{c}{\textbf{M3-Bench-web}} \\
\cmidrule(lr){2-4} \cmidrule(lr){5-7}
& Avg-R & R-Time & T-Time 
& Avg-R & R-Time & T-Time  \\
\midrule
\rowcolor{gray!7}
Baseline & 3.10 & 0.32 & 0.99 & 2.32 & 0.31 & 0.72 \\
\rowcolor{green!1}
Baseline + MemForest + AGPR & 2.52 & 0.32 & 0.81 & 2.12 & 0.32 & 0.68 \\
\rowcolor{green!4}
Baseline + MemForest + AGPR ( $\downarrow 30\%$ ) & 2.60 & 0.21 & 0.55 & 2.03 & 0.18 & 0.37 \\
\rowcolor{green!7}
Baseline + MemForest + AGPR ( $\downarrow 50\%$ ) & 2.59 & 0.16 & 0.41 & 2.20 & 0.16 & 0.35 \\
\rowcolor{green!10}
Baseline + MemForest + AGPR ( $\downarrow 70\%$ ) & 2.62 & 0.12 & 0.31 & 2.02 & 0.12 & 0.24 \\
\bottomrule[1.2pt]
\end{tabular}
\label{table14}
\end{table}
\vspace{-3pt}

Tables~\ref{table13}–\ref{table14} present the detailed results in Fig.~\ref{Fig3}, showing that our method achieves significant speedup.

\section{Case Study}
\label{appendixe}

\paragraph{EventTree Partitioning Process.}

The following shows an initial EventTree obtained during the EventTree partitioning process on the LoCoMo benchmark. The memory nodes within this EventTree primarily revolve around \textit{"transgender and community support"}. It can be observed that some memory nodes, though not temporally adjacent, exhibit high semantic similarity (e.g., nodes 1, 9, and 10, \textit{"related to community support"}), while other nodes, which are temporally adjacent but have lower semantic similarity with most memory nodes (e.g., nodes 7 and 8, \textit{"related to mentoring in the community"}), are also grouped into the same EventTree. This effectively integrates the global semantic similarity of memory events with their local temporal continuity.

\begin{tcolorbox}[
    colback=gray!5!white, 
    colframe=black!50, 
    title=An Example of an Initial EventTree Obtained from Partitioning, 
    sharp corners=southwest, 
    fonttitle=\bfseries,
    breakable 
]
\baselineskip=1.60em
\textbf{\textit{Memory 1: }}\textit{Values inclusivity and support.}\\
\textbf{\textit{Timestamp 1: }}\textit{1:14 pm on 25 May, 2023}\\\\
\textbf{\textit{Memory 2: }}\textit{Finds participating in charity events rewarding.}\\
\textbf{\textit{Timestamp 2: }}\textit{1:14 pm on 25 May, 2023}\\\\
\textbf{\textit{Memory 3: }}\textit{Aims to build a strong, supportive community of hope.}\\
\textbf{\textit{Timestamp 3: }}\textit{7:55 pm on 9 June, 2023}\\\\
\textbf{\textit{Memory 4: }}\textit{Believes in building a more inclusive and understanding world.}\\
\textbf{\textit{Timestamp 4: }}\textit{7:55 pm on 9 June, 2023}\\\\
\textbf{\textit{Memory 5: }}\textit{Excited to meet other people in the community.}\\
\textbf{\textit{Timestamp 5: }}\textit{1:36 pm on 3 July, 2023}\\\\
\textbf{\textit{Memory 6: }}\textit{Values the importance of fighting for trans rights and spreading awareness.}\\
\textbf{\textit{Timestamp 6: }}\textit{4:33 pm on 12 July, 2023}\\\\
\textbf{\textit{Memory 7: }}\textit{Has a mentee.}\\
\textbf{\textit{Timestamp 7: }}\textit{2:31 pm on 17 July, 2023}\\\\
\textbf{\textit{Memory 8: }}\textit{Mentors a transgender teen.}\\
\textbf{\textit{Timestamp 8: }}\textit{2:31 pm on 17 July, 2023}\\\\
\textbf{\textit{Memory 9: }}\textit{Passionate about rights and community support.}\\
\textbf{\textit{Timestamp 9: }}\textit{8:56 pm on 20 July, 2023}\\\\
\textbf{\textit{Memory 10: }}\textit{User is passionate about rights and community support.}\\
\textbf{\textit{Timestamp 10: }}\textit{8:56 pm on 20 July, 2023}\\\\
\textbf{\textit{Memory 11: }}\textit{The group has regular meetings and plans events and campaigns.}\\
\textbf{\textit{Timestamp 11: }}\textit{8:56 pm on 20 July, 2023}\\\\
\textbf{\textit{Memory 12: }}\textit{Passionate about helping people and making a positive impact.}\\
\textbf{\textit{Timestamp 12: }}\textit{8:56 pm on 20 July, 2023}\\\\
\textbf{\textit{Memory 13: }}\textit{Believes in the fight for equality and inclusivity.}\\
\textbf{\textit{Timestamp 13: }}\textit{8:56 pm on 20 July, 2023}\\\\
\textbf{\textit{Memory 14: }}\textit{Values standing up for equality.}\\
\textbf{\textit{Timestamp 14: }}\textit{2:24 pm on 14 August, 2023}\\\\
\textbf{\textit{Memory 15: }}\textit{Wants to spread understanding and acceptance.}\\
\textbf{\textit{Timestamp 15: }}\textit{1:33 pm on 25 August, 2023}\\\\
\textbf{\textit{Memory 16: }}\textit{Shared personal story with young people.}\\
\textbf{\textit{Timestamp 16: }}\textit{3:19 pm on 28 August, 2023}\\\\
\textbf{\textit{Memory 17: }}\textit{Proud of her identity.}\\
\textbf{\textit{Timestamp 17: }}\textit{12:09 am on 13 September, 2023}
\end{tcolorbox}

\paragraph{Memory Node Merging Process.}

As shown below, we present an example from the M3-Bench-robot benchmark during the memory node merging process. The two nodes to be merged are highly related, and the resulting merged memory node fully preserves the original information while significantly improving storage efficiency and enhancing subsequent retrieval efficiency.

\begin{tcolorbox}[
    colback=gray!5!white, 
    colframe=black!50, 
    title=An Example of Merging Two Memory Nodes, 
    sharp corners=southwest, 
    fonttitle=\bfseries,
    breakable 
]
\baselineskip=1.60em
\textbf{\textit{Memory 1: }}\textit{The task involves organizing and preparing documents, possibly for a presentation or application.}\\
\textbf{\textit{Timestamp 1: }}\textit{Clip 6}\\\\
\textbf{\textit{Memory 2: }}\textit{The task being performed involves document preparation, possibly for an application, report, or presentation.}\\
\textbf{\textit{Timestamp 2: }}\textit{Clip 6}\\\\
\textbf{\textit{Merged memory: }}\textit{The task involves organizing and preparing documents for a presentation, application, or report.}\\
\textbf{\textit{Timestamp: }}\textit{Clip 6}
\end{tcolorbox}

\section{Limitations}

We conducted experiments on both the high-redundancy multimodal memory framework M3-Agent and the low-redundancy textual memory framework Mem0. The results indicate that in Mem0, which has relatively low redundancy, performance degradation becomes pronounced at higher compression rates. For instance, when \textbf{70\%} of the historical memory is compressed, the accuracy drops to only \textbf{93.3\%} of that in the uncompressed setting. In the future, specialized compression methods could be designed for such low-redundancy memory frameworks to improve performance under high compression ratios.

\section{Future Work}
\label{appendixg}

We propose the following potential research directions for future work:

\begin{itemize}[leftmargin=1em, itemsep=0pt, topsep=0pt]
    \item MemForest is a general-purpose memory compression framework. Future work could explore designing specialized compression methods for low-redundancy memory frameworks like Mem0, in order to better preserve performance under high compression rates.
    \item As discussed in Section~\ref{appendixd.6}, it may be possible to train a dedicated small memory-merging model that achieves, or even surpasses, the memory node merging performance of large models such as GPT-4o-mini.
    \item There is also room for optimization in memory retrieval mechanisms. Most current methods rely on selecting the top \( k \) most similar memory nodes, which involves only limited embedding similarity computations and can leverage vector database acceleration, typically resulting in retrieval times under one second. In contrast, some newer approaches, such as xMemory \cite{r43}, require redesigning the memory generation framework, are incompatible with existing frameworks, and perform extensive embedding similarity computations during retrieval. They also require reconsideration of memory structure and cannot leverage vector database acceleration, leading to retrieval times that are tens of times longer. Therefore, future work could explore retrieval mechanisms that minimize embedding similarity computations while achieving a better balance between retrieval speed and accuracy.
    \item In terms of safety, it may be possible to perform a holistic analysis of the historical memory set to identify conflicts or potentially corrupted entries, and take appropriate measures to correct or remove them, thereby enhancing the robustness and reliability of the memory system.
\end{itemize}



\end{document}